\documentclass[sigconf, 10pt]{acmart}
\newif\ifnotnewacm
\notnewacmfalse

\newif\ifheadnice
\headnicefalse

\begin{CJK}{UTF8}{gbsn}

\newtheorem{algorithm}{算法}

\renewcommand{\abstractname}{摘要}  
\renewcommand{\refname}{参考文献}   

\title{***}
\author{***footnote{电子邮件:**}\\[2ex]
*** \\[2ex]
}
\date{20XX年X月}

\maketitle

\tableofcontents
\newpage
在此输入正文，中英文均可。

\songti 
\fangsong 
\biaosong
\heiti
\kaishu

\end{CJK}

\newif\ifeg
\egfalse

\renewcommand\footnotetextcopyrightpermission[1]{} 
\def\pku{\superscript{$\mathbb{P}$}}

\def\uestc{\superscript{$\mathbb{U}$}}
\def\hcs{\superscript{$\mathbb{K}$}}
\def\mit{\superscript{$\mathbb{M}$}}
\def\pitt{\superscript{$\mathbb{PI}$}}

\usepackage{setspace}
\usepackage{adjustbox}
\usepackage{tcolorbox}
\usepackage{makecell}

\def\sys{\textsc{WiCompass}\xspace}

\newif\ifshowrevisions
\showrevisionsfalse

\ifshowrevisions
  \newcommand{\revision}[1]{\textcolor{blue}{#1}}
\else
  \newcommand{\revision}[1]{#1}
\fi

\newcommand{\revtag}[1]{%
  \ifshowrevisions
    \textbf{#1}%
  \fi
}

\begin{document}

\title{\sys: Oracle-driven Data Scaling for \\ mmWave Human Pose Estimation}

\author{Bo Liang\pku, Chen Gong\pku, Haobo Wang\pku, Qirui Liu\pku, Rungui Zhou\pku, Fengzhi Shao\uestc, Yubo Wang\uestc, Wei Gao\pitt, Kaichen Zhou\mit, Guolong Cui\uestc, Chenren Xu\pku\hcs\superscript{\Letter}}

\authornote{
    \Letter: chenren@pku.edu.cn
}

\affiliation{
    \vspace{-3mm}
    \begin{tabular}{c}
        {\pku}School of Computer Science, Peking University {\uestc}University of Electronic Science and Technology of China \\
        {\pitt}University of Pittsburgh {\mit}Electrical Engineering and Computer Science, Massachusetts Institute of Technology \\
        {\hcs}Key Laboratory of High Confidence Software Technologies, Ministry of Education (PKU)
    \end{tabular}
    \vspace{1mm}
}

\def\authors{Bo Liang, Chen Gong, Haobo Wang, Qirui Liu, Rungui Zhou, Fengzhi Shao, Yubo Wang, Kaichen Zhou, Wei Gao, Guolong Cui, Chenren Xu}

\begin{abstract}
Millimeter-wave Human Pose Estimation (mmWave HPE) promises privacy but suffers from poor generalization under distribution shifts. We demonstrate that brute-force data scaling is ineffective for out-of-distribution (OOD) robustness; efficiency and coverage are the true bottlenecks. To address this, we introduce \sys, a coverage-aware data-collection framework. \sys leverages large-scale motion-capture corpora to build a universal pose space ``oracle'' that quantifies dataset redundancy and identifies underrepresented motions. Guided by this oracle, \sys employs a closed-loop policy to prioritize collecting informative missing samples. Experiments show that \sys consistently improves OOD accuracy at matched budgets and exhibits superior scaling behavior compared to conventional collection strategies. By shifting focus from brute-force scaling to coverage-aware data acquisition, this work offers a practical path toward robust mmWave sensing.
\end{abstract}



\acmYear{2026}\copyrightyear{2026}
\acmConference[MobiCom '26]{The 32nd Annual International Conference on Mobile Computing and Networking}{October 26--30, 2026}{Austin, TX, USA}
\acmBooktitle{The 32nd Annual International Conference on Mobile Computing and Networking (MobiCom '26), October 26--30, 2026, Austin, TX, USA}
\acmDOI{10.1145/3795866.3796684}
\acmISBN{979-8-4007-2505-0/26/10}

\maketitle

\section{Introduction}\label{sec:intro}

Human Pose Estimation (HPE)—the task of localizing anatomical keypoints to recover body posture~\cite{zheng2023deep}—is a cornerstone technology for applications in human–computer interaction~\cite{hidalgo2019single}, automated healthcare~\cite{swordhealth}, and immersive media~\cite{google_mediapipe_pose_misc}. The dominant approach relies on optical sensors such as RGB cameras which, despite high accuracy, suffer fundamental limitations: they compromise privacy, perform poorly under inconsistent lighting, and are susceptible to occlusion. In response, millimeter-wave (mmWave) radar has emerged as a compelling alternative. By nature, mmWave sensing is privacy-preserving, robust to lighting changes, and capable of penetrating many non-metallic materials~\cite{sengupta2020mm, shi2022mpose, xue2023towards, hu2024mmpose}.

Despite this promise, progress in mmWave HPE rests on a fragile data foundation. While state-of-the-art (SOTA) models achieve impressive in-distribution (ID) joint localization errors comparable to optical HPE~\cite{xue2021mmmesh, kong2022m3track, mmdiff, Chen2024XFiAM}, their performance degrades sharply under even modest data distribution shifts. This brittleness is stark in our leave-one-out test (\secref{ssec:bottleneck}): simply withholding a single action such as ``Waving hand (left)'' from the training set causes the human joint localization error on that unseen action to spike from roughly 50\,mm to well over 120\,mm. The disproportionate failure, even when semantically related motions (``Waving hand (right)'' and ``Raise hand (left)'') remain in the training set, points to a critical problem: current models learn to interpolate expertly within a known distribution but fail to generalize across the true, diverse spectrum of human motion. This gap between performance in the lab and reliability in the wild is a critical barrier to deployment.

The intuitive response -- simply scaling up data via additional real-world collection or synthetic generation -- is a false panacea. Real-world acquisition is notoriously expensive in time, personnel, and setup costs~\cite{wang2024xrf55,mmfi,chen2022mmbody,ho2024rt}. Achieving coverage comparable to vision HPE datasets~\cite{ionescu2013human3,Joo_2015_ICCV,cai2022humman} typically requires recruiting tens to hundreds of participants and capturing across multiple environments, with calendar time measured in months and total human-hours in the thousands. While data synthesis with simulators~\cite{yu2022rfgan,chen2023rf,chi2024rf} can be faster, it often suffers a persistent ``sim-to-real'' gap, failing to capture the nuances of real-world physics and misaligning with in-the-wild data distributions~\cite{gong2025data}. More critically, our analysis shows that naively adding more data is not only inefficient but frequently ineffective. A progressive subset study (\secref{ssec:bottleneck}) on existing large-scale datasets reveals this redundancy: randomly discarding up to 70\% of training samples of existing datasets results in less than a 2\% drop in performance. These findings shift the focus from \textit{data quantity} to \textit{data efficiency}, and leads to the central question of this paper: \textit{What is the optimal data acquisition strategy to maximize the generalization of mmWave HPE models?}


We address this challenge with \sys, an oracle-driven data collection framework for mmWave HPE, where the ``oracle'' is a high-coverage geometric prior over human poses. We distill motion priors from large-scale motion-capture (MoCap) corpora into a shared latent pose space. By projecting pose labels (skeletons) from both mmWave samples and reference MoCap into this space, we quantify coverage and redundancy via $k$-Nearest Neighbors ($k$-NN), identifying underrepresented poses and oversampled regions. We then convert these diagnostics into a closed-loop acquisition strategy that prioritizes informative targets—via real-world capture or simulation—under a fixed collection budget. Importantly, \sys does not attempt to align raw mmWave measurements with optical signals; it operates on modality-agnostic pose labels, so coverage analysis is not contingent on cross-modal feature alignment. \textit{\sys is named to evoke a ``compass'' for data collection: it provides a principled guideline that steers mmWave HPE toward the most informative motions, rather than brute-force data scaling.} Our contributions are threefold:

\begin{itemize}[leftmargin=10pt]
\item \textbf{Empirical diagnosis of the coverage bottleneck.} We identify insufficient data coverage—not model capacity—as the primary bottleneck in mmWave HPE (\secref{sec:motivation}). Our analysis reveals severe out-of-distribution (OOD) fragility and extensive redundancy in current datasets.

\item \textbf{An interpretable data coverage framework.} We propose a principled, backbone-independent framework to quantify coverage and redundancy in a latent pose space (\secref{sec:vqvae_tokenizer} and \secref{sec:knn_framework}). It provides visualizable diagnostics that pinpoint concrete gaps in existing data.

\item \textbf{An actionable and coverage-efficient acquisition strategy.} Guided by these diagnostics, we design \sys, a closed-loop acquisition approach that targets underrepresented poses (\secref{sec:data_collection}). It delivers superior generalization with orders of magnitude fewer samples than baselines.
\end{itemize}

\textit{This study does not raise ethical concerns. All the code, datasets, and experiment scripts are publicly available at} \\
\textcolor{blue}{\texttt{https://github.com/Galaxywalk/WiCompass}}
\section{Pilot Study and Motivation}\label{sec:motivation}
\begin{figure*}[t]
    \centering
    \begin{subfigure}[b]{0.32\textwidth}
        \centering
        \includegraphics[width=\linewidth]{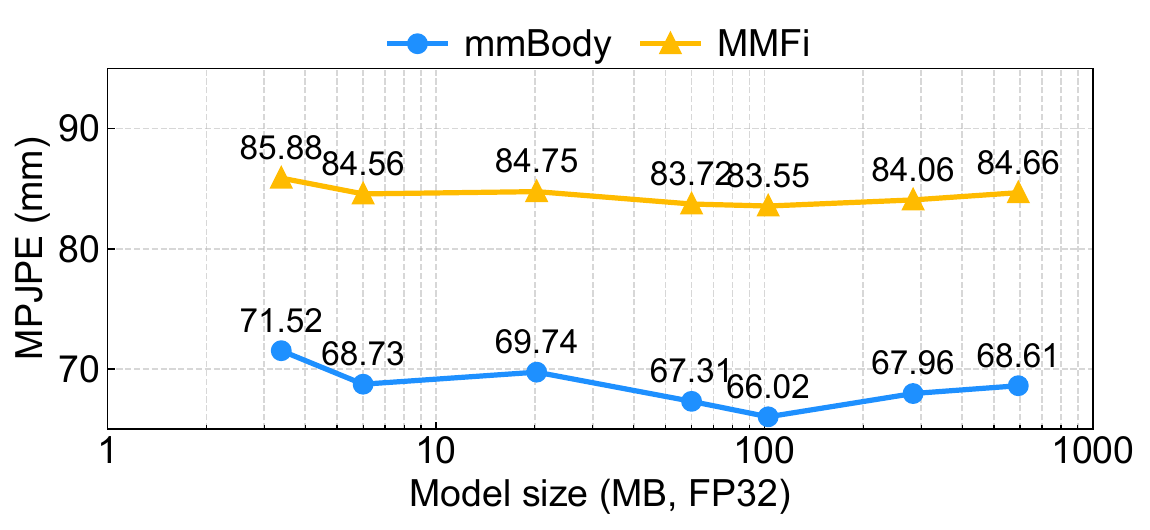}
        \caption{Model Performance vs. Model Size.}
        \label{fig:size_vs_mpjpe}
    \end{subfigure}\hfill
    \begin{subfigure}[b]{0.32\textwidth}
        \centering
        \includegraphics[width=\linewidth]{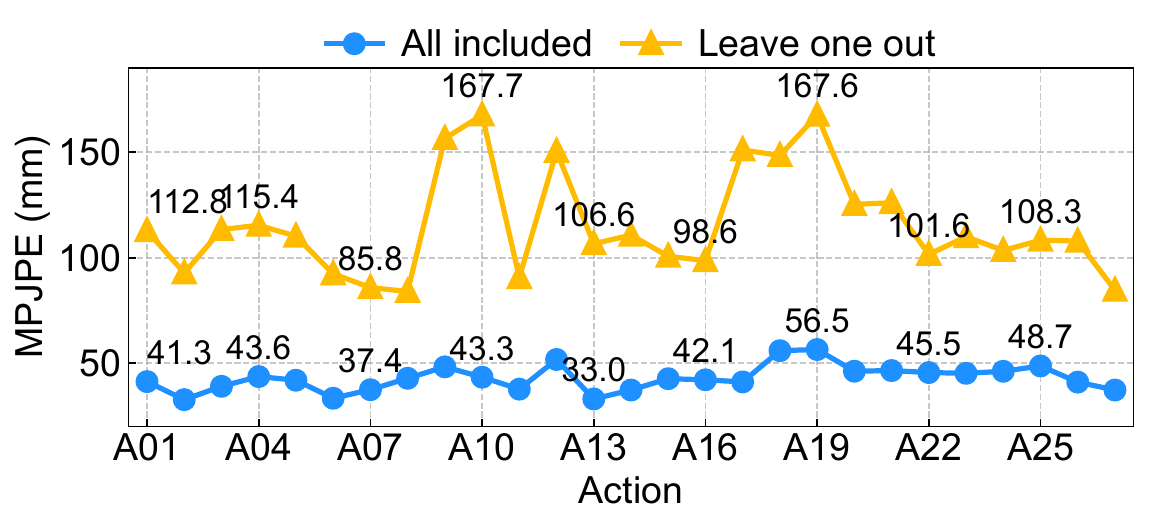}
        \caption{Generalization Tests.}
        \label{fig:leave_one_out_generalization}
    \end{subfigure}\hfill
    \begin{subfigure}[b]{0.32\textwidth}
        \centering
        \includegraphics[width=\linewidth]{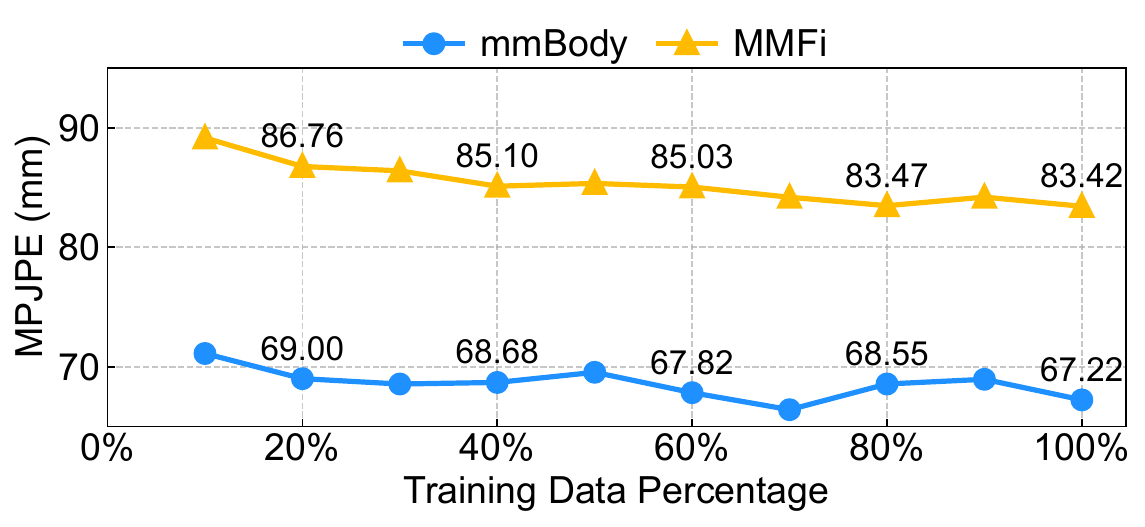}
        \caption{Dataset Efficiency.}
        \label{fig:action_generalization}
    \end{subfigure}
    \vspace{-2mm}
    \caption{An overview of the pilot studies. (a) illustrates the relationship between model performance and size. (b) shows the results of leave-one-out generalization tests. (c) demonstrates the data efficiency of existing datasets.}
    \label{fig:pilot_studies}
\end{figure*}

This section probes the main obstacles to progress in mmWave HPE. We first define the task, then present a pilot study showing that the dominant constraints are \textit{data generalization} and \textit{data efficiency}, rather than \textit{model architectural capacity}.

\begin{table}[ht]
\centering
\caption{Open Source mmWave HPE Datasets.}
\vspace{-3mm}
\label{tab:datasets}
\resizebox{\linewidth}{!}{%
\begin{tabular}{lllll}
\toprule
Datasets & Data Format & Label Style & Size  & Extra Annotations            \\ \midrule
MRI~\cite{an2022mri}   & Point Cloud (PD)      &      COCO      &160k       & Subj, Scene, Action   \\ 
mmBody~\cite{chen2022mmbody}   & PD & SMPL-X      & 67k        & Subj, Scene   \\
MMFi~\cite{mmfi}     & PD & Human3.6M   & 321k    & Subj, Scene, Action         \\
MilliPoint~\cite{cui2023milipoint}   & PD    & Customized            & 213k      & Subj, Action                            \\
HuPR~\cite{lee2023hupr}     & Heatmap (HP)     & LSPe             & 36k        &  No                         \\
RT-Pose~\cite{ho2024rt}  & HP \& PD & Human3.6M             & 72k            & (Multi-)Subj, Scene, Action \\
\bottomrule
\end{tabular}%
}
\vspace{-3mm}
\end{table}

\subsection{Task Definition}
The objective of this work is to perform 3D HPE from mmWave radar data. We define the task as follows:

\nosection{Input and Output} The deep learning model takes a mmWave point cloud of a human subject as \textit{input} and regresses the 3D coordinates of the body joints as \textit{output}. The input point cloud is generated from raw FMCW signals via a standard processing pipeline, which includes FFTs to create a radar heatmap and a Constant False Alarm Rate algorithm (CFAR)~\cite{zhang2023lt} for point extraction. For the output, we adopt the SMPL-X skeleton format~\cite{pavlakos2019expressive} of 22 main joints, chosen for its compatibility with motion capture data over other common formats like COCO~\cite{lin2014microsoft} and Human3.6M~\cite{ionescu2013human3}.

\nosection{Datasets}
An overview of publicly available mmWave HPE datasets is provided in ~\tabref{tab:datasets}. From this pool, we selected the mmBody~\cite{chen2022mmbody} and MMFi~\cite{mmfi} datasets for our experiments. Other datasets were intentionally excluded due to issues that would compromise experimental validity, such as poor data quality (e.g., misalignment between sensor readings and ground-truth labels) or the use of custom data formats incompatible with our processing pipeline.

\nosection{Metrics}
Our primary evaluation metric is the Mean Per-Joint Position Error (MPJPE), reflecting a model's absolute spatial accuracy. We relegate another common metric, Procrustes Aligned MPJPE (PA-MPJPE) to a secondary role. PA-MPJPE applies a rigid transformation (rotation, translation, scaling) to align poses before error calculation -- a justifiable step for vision-based methods contending with 2D-to-3D scale ambiguities. This rationale is inapplicable to mmWave sensors, which provide direct and unambiguous 3D metric data. For this modality, any spatial error is a direct model failure, not a sensor artifact. PA-MPJPE's alignment masks these critical failures, yielding an overly optimistic score unrepresentative of real-world utility. We, therefore, contend that MPJPE is the truer measure of a deployable system's performance, and use PA-MPJPE exclusively as a diagnostic tool during model training.

\begin{table}
\centering
\caption{Comparison of MPJPE Errors Across Datasets and Models.}
\vspace{-3mm}
\label{tab:benchmark_results}
\resizebox{\linewidth}{!}{%
\begin{tabular}{lcccc}
\toprule
& Original paper & X-Fi~\cite{Chen2024XFiAM} & mmDiff~\cite{mmdiff}  & \textbf{Our Model} \\
\midrule
mmBody~\cite{chen2022mmbody} & 82.00 & - & 68.08 & \textbf{67.22}\\
MMFi~\cite{mmfi} & 136.1 & 127.4 & \textbf{82.73} & 83.42   \\
\bottomrule
\end{tabular}%
}
\end{table}

\subsection{Model is Not the Bottleneck}\label{ssec:model_pilot_study}
Motivated by scaling-law thinking in deep learning, we ask whether progress in mmWave HPE comes mainly from larger or more sophisticated models, or whether the true bottleneck lies elsewhere~\cite{hestness2017deep, kaplan2020scaling, zhai2022scaling}. To test this, we use a scalable Point-Transformer model~\cite{zhao2021point} with a transformer decoder and learnable joint queries. We vary model capacity by adjusting depth and width and train all variants on the mmBody and MMFi datasets (details in Appendix~\ref{app:model}) under an identical protocol. The trend in \figref{fig:size_vs_mpjpe} is clear: error decreases at first but quickly plateaus around a model size of $\sim$10\,MB, after which adding more parameters gives only small gains or even worse performance. Compute is not a limiting factor, so this saturation cannot be explained by training budget. We then compare our baseline—the 102.64 M-parameter variant in \figref{fig:size_vs_mpjpe}, also used in later experiments—against SOTA methods. As shown in \tabref{tab:benchmark_results}, this straightforward architecture matches or even outperforms more complex designs on both MMFi and mmBody (\eg X-Fi~\cite{Chen2024XFiAM} uses multi-modal pretraining, while mmDiff~\cite{mmdiff} adds physical and pose priors). Together, these results show that at the current data scale, model capacity—and by extension, compute—is not the limiting factor. This agrees with established scaling-law results highlighting the joint roles of model, data, and compute~\cite{kaplan2020scaling, zhai2022scaling, li2025mis}. We therefore turn our focus to data coverage and redundancy, which we identify as the true bottlenecks for generalization in mmWave HPE.

\subsection{The Dual Data Bottlenecks}\label{ssec:bottleneck}

Our analysis uncovers two data-level issues in current mmWave datasets: (i) poor out-of-distribution (OOD) generalization on unseen actions, and (ii) low data efficiency due to heavy redundancy.

\nosection{Generalization Bottleneck}
Although ID metrics for mmWave HPE look strong (see \tabref{tab:benchmark_results}, even comparable to vision-based methods), models fail badly on unseen variations and tend to overfit the training distribution. To evaluate this systematically, we conduct a \textit{leave-one-out} (LOO) action split on the MMFi dataset~\cite{mmfi}~\footnote{MMFi provides annotations for 40 subjects, 27 actions, and 4 environments. We exclude mmBody because it lacks action annotations.}. In each round, the model trains on 26 actions and is tested on the held-out one, repeating this process for all 27 actions with the same training recipe as \secref{ssec:model_pilot_study}. For comparison, we also train an \textit{all-included} (ID) model using all training splits. The results in \figref{fig:leave_one_out_generalization} reveal several issues:

\begin{itemize}[leftmargin=10pt]
    \item \textbf{Severe overfitting from data sparsity.} While the ID MPJPE is \(50.64\pm7.67\) mm,\footnote{The original dataset split is both cross-scene and cross-subject in \tabref{tab:benchmark_results}, so the testing error decreases from 82.42 mm to 50.64 mm in this fully ID case.} the error on these held-out actions jumps to \(122.10\pm26.76\) mm ($\sim 2.4\times$). This shows that the dataset lacks sufficient diversity to support generalization across unseen actions.
    \item \textbf{Insufficient variation prevents learning symmetries.} Failures on symmetric action pairs (e.g., ``Lunge (toward left-front/right-front)'', A9/A10; ``Waving hand (left/right)'', A17/A18) indicate that mirrored counterparts are treated as unrelated. Although both appear in the dataset, the model fails to learn the underlying symmetry, but merely employs a brittle memorization.
    \item \textbf{Outlier poses highlight imbalance.} Disproportionately high errors on complex actions such as “Picking up things” (A19) expose structural imbalance: the dataset is dominated by common upright postures, leaving rare but important motions underrepresented. Apparent wins (e.g., “Bowing,” A27) occur because key torso configurations are incidentally covered by another action (“Picking up things,” A19), rather than from a robust representation of motion space. Such reliance on chance overlaps reflects insufficient diversity, producing models whose success is fragile and inconsistent.
\end{itemize}

\nosection{Efficiency Bottleneck}
A natural response to weak generalization is to scale up data, yet current collection practices are inefficient. We run a \textit{progressive-subset} test—train the same model on random subsets of the training split of sizes of $100\%, 90\%, \dots, 10\%$, then evaluate on the original test set.
As shown in \figref{fig:action_generalization}, performance stays almost unchanged until the subset shrinks below $\sim\!30\%$: discarding about $70\%$ of the training samples only increases error by <2\%. This is not robustness but redundancy. Human-intuitive collection (e.g., pre-planned action lists, repeated takes of the same motion) oversamples already-dense regions of pose space while leaving important regions sparse. A large share of effort thus goes to near-duplicates that add little new information.

\begin{figure*}[t]
    \centering
    \includegraphics[width=\linewidth]{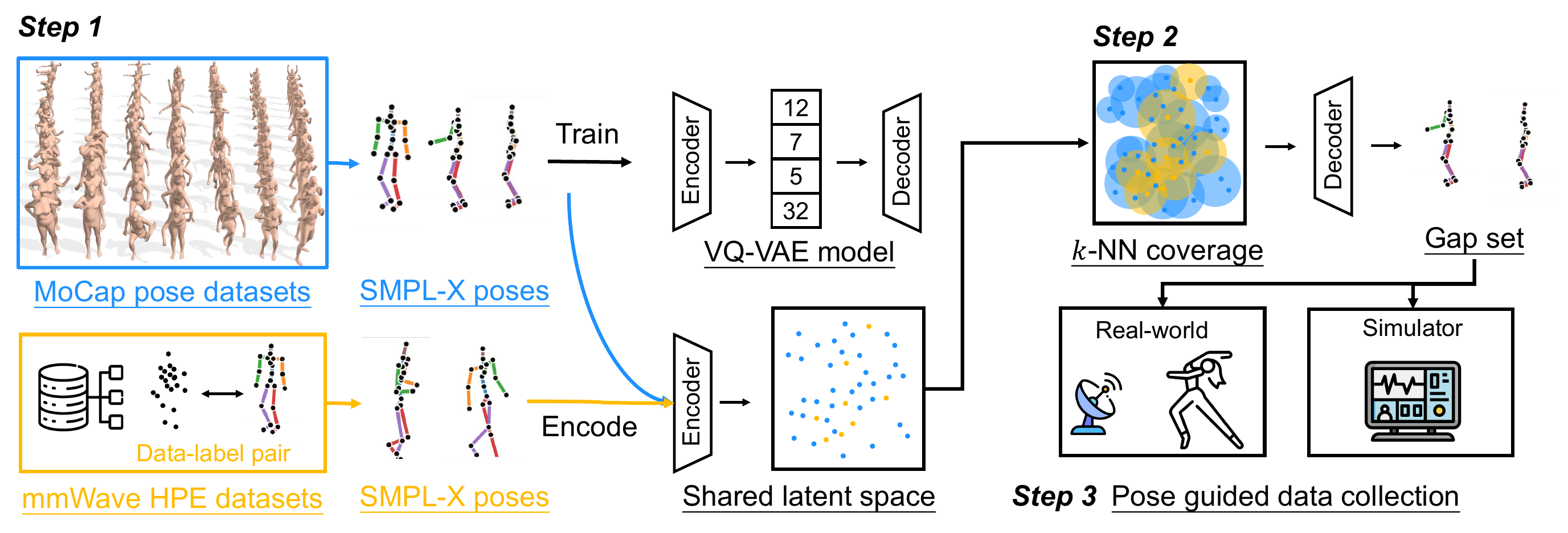}
    \vspace{-4mm}
    \caption{\sys Overview. Human poses of SMPL-X format from MoCap datasets and mmWave datasets are encoded into a shared latent space via VQ-VAE. Directional $k$-NN coverage identifies uncovered regions, which then guides pose-conditioned data collection in real or simulated settings.} 
    \label{fig:overview}
\end{figure*}

\subsection{Summary and Motivation}
Our pilot study rules out a model-centric explanation and identifies data as the primary obstacle to robust mmWave HPE. Two challenges emerge clearly:

\begin{itemize}[leftmargin=10pt]
    \item \textbf{Generalization Bottleneck.} Models do not learn invariant motion representations and can treat even mirrored actions as OOD, a gap that vision systems often close with massive, diverse datasets that implicitly teach spatial equivariances. It shows that mmWave datasets have limited coverage to various human motion data.
    \item \textbf{Efficiency Bottleneck.} Naively “scaling up” data under current practices yields highly redundant collections, consuming budget on near-duplicates while failing to cover informative, underrepresented motions.
\end{itemize}

This creates a paradox: we need substantially more diverse data, but existing pipelines acquire it inefficiently. To resolve this, we introduce \sys, an ``oracle-driven'' framework that aligns to a comprehensive pose latent space and uses quantitative coverage analysis to target collection toward maximally informative gaps—improving generalization without brute-force scaling.
\section{Overview}
\label{sec:overview}

The central challenge in mmWave HPE lies in the prohibitive cost of building datasets that are both large and truly diverse. Existing mmWave corpora remain narrow in scope, whereas the motion-capture community has curated broad repositories such as AMASS~\cite{mahmood2019amass}, which span a much richer spectrum of human motion. Our key idea is to treat this motion prior as an ``oracle'' that guides which mmWave samples should be collected next: instead of indiscriminately expanding data volume, we selectively target the samples that contribute most to generalization.

\figref{fig:overview} depicts our framework, \sys, which operationalizes this principle through a compact, cross-modal pipeline with three stages.  
\textit{(1) Shared latent space.} We begin by training a Vector Quantized Variational Autoencoder (VQ-VAE) model on AMASS to construct a modality-agnostic vocabulary of pose primitives. Both MoCap and mmWave poses are then embedded into this shared latent space (\secref{sec:vqvae_tokenizer}).  
\textit{(2) Coverage analysis.} Within this space, we conduct a directional $k$-NN analysis to measure how much of the MoCap manifold is represented by the mmWave dataset and to reveal underrepresented regions. This analysis produces a multi-scale \emph{coverage map}, and we further introduce a scale-free redundancy metric to quantify oversampling (\secref{sec:knn_framework}). 
\textit{(3) Guided acquisition.} Based on these diagnostics, \sys selects a batch of targets under a fixed budget using capped probability-proportional-to-size sampling: extreme outliers are first filtered out, after which candidate poses are sampled in proportion to their coverage density. The mmWave data for these targeted poses are then obtained via a unified interface supporting both real-world and simulation, and the resulting samples are merged back into the dataset (\secref{sec:data_collection}).

Together, these stages form a closed feedback loop: \sys identifies coverage gaps, translates them into concrete acquisition targets, and iteratively enriches dataset diversity while improving generalization efficiency.
\section{Learning a Universal Vocabulary of Human Motion}
\label{sec:vqvae_tokenizer}

Our pilot study highlights a critical bottleneck: mmWave HPE models suffer from poor generalization and redundancy. To overcome this, we advocate a shift in data scaling—from brute-force collection to a targeted, informed strategy. The key insight is that while existing mmWave datasets are sparse, the manifold of human motion is already comprehensively captured by large-scale MoCap datasets such as AMASS~\cite{mahmood2019amass}. We can transfer this rich motion knowledge from the MoCap domain into the wireless domain.

This requires a common and high accuracy representation of human motion. Directly using raw 3D joint coordinates (66-D for 22 joints) is ill-suited: high dimensionality makes distances unreliable indicators of perceptual or semantic similarity between poses, and the space itself is unstructured and redundant. Instead, we learn a compact latent representation that encodes poses into a semantically meaningful form. In this section, we describe how we construct such a representation by training a ``universal pose tokenizer'' on large-scale MoCap data.

\begin{figure}[h]
    \centering
    \includegraphics[width=\linewidth]{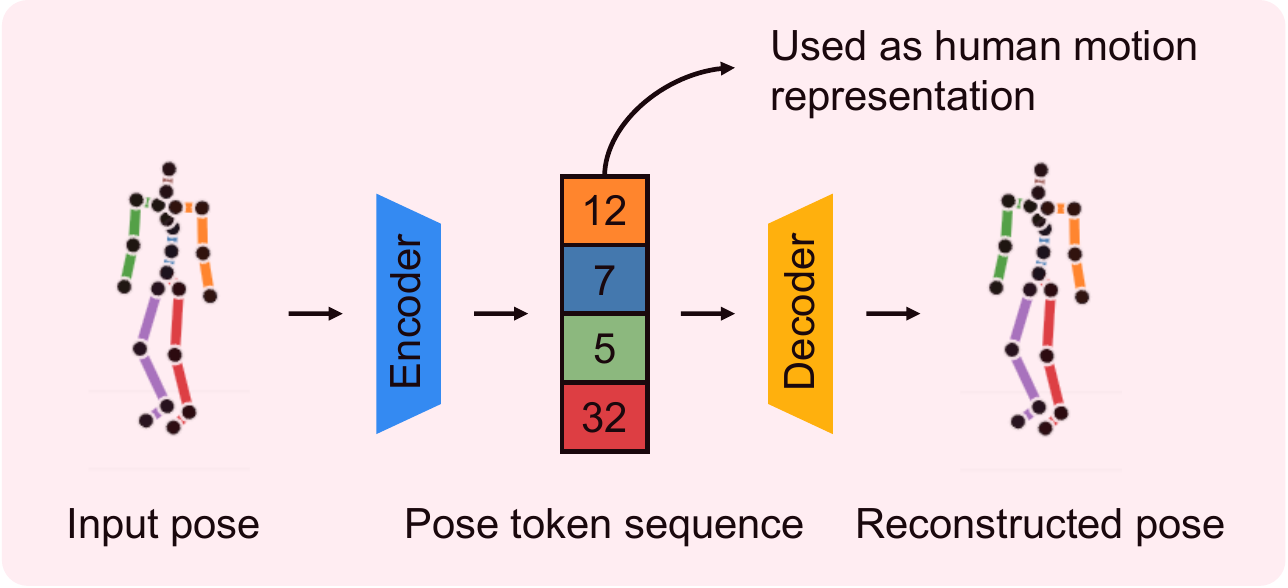}
    \vspace{-4mm}
    \caption{VQ-VAE Model Architecture.}
    \vspace{-3mm}
    \label{fig:vqvae}
\end{figure}

\subsection{A Discrete and Compositional Pose Representation}

We employ a standard VQ-VAE architecture~\cite{van2017neural}, trained exclusively on MoCap data, to learn a universal motion vocabulary (\figref{fig:vqvae}). A VQ-VAE is an autoencoder that compresses input poses into a latent space and reconstructs them from this compressed representation. Its distinctiveness lies in the latent space: rather than being continuous, it consists of a finite \textit{codebook} of discrete \textit{tokens}. Each pose must be reconstructed as a composition of these tokens, forcing the model to discover recurring motion patterns and store them as ``words'' in the codebook. The result is a meaningful, learned pose vocabulary. Representing any pose as a \textit{token sequence} brings three key advantages:

\begin{itemize}[leftmargin=8pt]
    \item \textbf{Modality Agnosticism via Quantization.} Quantization maps any 3D pose—whether from optical MoCap or mmWave estimates—into a shared set of tokens. This yields a canonical vocabulary that is inherently modality-independent and enables direct comparison.

    \item \textbf{Tractable Distributional Analysis.} The discrete space makes coverage analysis straightforward: instead of comparing continuous distributions, we simply count tokens' distance. This renders the analysis computationally tractable, robust, and interpretable. Recent studies also show that discrete latent spaces enhance learning capacity and capture long-range motion dependencies~\cite{guo2022tm2t, zhang2023generating, saleem2025genhmr}.

    \item \textbf{Decoupling from Downstream Backbones.} Our latent space is learned through a generic reconstruction task, making it task-agnostic. In contrast, using features from a task-specific backbone (e.g., action classifiers) would introduce biases tied to that model’s architecture and objective.
\end{itemize}

\subsection{Model Architecture}
The model architecture is shown in \figref{fig:vqvae}, which consists of an encoder $E$, a discrete codebook $\mathcal{C} = \{\mathbf{c}_k\}_{k=1}^K \in \mathbb{R}^{K \times D}$, and a decoder $D$. The encoder maps an input 3D pose $\mathbf{x} \in \mathbb{R}^{J \times 3}$ to a sequence of continuous latent vectors $\mathbf{z}_e = E(\mathbf{x}) \in \mathbb{R}^{T \times D}$. Here, $J$ denotes the number of joints (in our case, $J = 22$), $T$ is the number of tokens (sequence length), and $D$ is the dimension of each latent embedding. The codebook contains $K$ distinct embedding vectors. The vector quantization mechanism, $Q$, then maps each continuous vector $z_{e,t}$ to its closest codebook embedding $\mathbf{c}_k$ via a nearest-neighbor lookup:
\begin{equation}
Q(z_{e,t}) = \mathbf{c}_k \quad \text{where} \quad k = \arg\min_{j \in \{1, \dots, K\}} ||z_{e,t} - \mathbf{c}_j||_2^2
\end{equation}
The decoder $D$ then attempts to reconstruct the original pose $\hat{\mathbf{x}} = D(Q(\mathbf{z}_e))$ from the sequence of quantized vectors.

\subsection{Loss and Training}
\revision{\revtag{(Meta.3)}
We train the model end-to-end on the AMASS dataset using the standard VQ-VAE objective, combining reconstruction loss with codebook commitment terms.} The composite loss function balances reconstruction fidelity with the stability of the discrete latent space:
\begin{equation}
    \mathcal{L} = \mathcal{L}_{\text{recon}} + ||\text{sg}(\mathbf{z}_e) - \mathbf{z}_q||_2^2 + \beta ||\mathbf{z}_e - \text{sg}(\mathbf{z}_q)||_2^2
\end{equation}
\revision{\revtag{(B.4)}
where $z_q = Q(z_e)$ is the \textit{quantized latent vector}, formed by replacing each continuous encoder output with its nearest codebook entry. The first term, $\mathcal{L}_{recon} = \|x - \hat{x}\|_2^2$, ensures reconstruction fidelity. The second term is VQ loss, which pulls codebook embeddings toward the encoder outputs, and the third term is the commitment loss (weighted by $\beta$), which encourages the encoder to commit to the discrete codebook vectors.} To further improve stability, we update the codebook embeddings using an Exponential Moving Average (EMA) of the encoder outputs, which maintains a running average of the codebook embeddings and provides a more stable update rule compared to direct gradient-based updates \cite{razavi2019generating}.

Upon completion of training, the frozen encoder and codebook serve as our universal pose tokenizer. For any pose---whether from the reference optical MoCap dataset or a mmWave dataset---we can generate a canonical feature vector $\phi$ by looking up its token sequence in the codebook and concatenating the embeddings. This process projects poses from disparate sources into a single, unified latent space, enabling the direct distributional comparison detailed in ~\secref{sec:knn_framework}. The decoder, however, is retained; it plays a crucial role in \secref{sec:data_collection}, where we use it to visualize the identified coverage gaps, allowing us to generate and inspect concrete examples of the missing poses that need to be collected.

\section{Dataset Coverage Analysis}
\label{sec:knn_framework}

Having established a universal motion vocabulary in the latent space (\secref{sec:vqvae_tokenizer}), we now introduce a framework to quantify how well a given dataset occupies this space. Our goal is to move beyond anecdotal statements such as ``this dataset lacks sitting motions'' or ``there is a lot of redundancy'' and instead define metrics that: (i) measure how much of the motion manifold a dataset actually covers, and (ii) reveal how efficiently it uses its samples.

We build a $k$-Nearest Neighbors ($k$-NN)–based framework that provides a unified view of dataset quality from two complementary perspectives:
(i) a \textit{cross-dataset} analysis that identifies what is missing by comparing a mmWave dataset against a larger MoCap corpus, and
(ii) an \textit{intra-dataset} analysis that characterizes redundancy and internal coverage. In this subsection, we first define local neighborhoods in the latent manifold; later, we instantiate these neighborhoods into concrete coverage metrics.

\subsection{Local Neighborhoods via k-NN}
\label{sec:knn_metric}

Our approach is inspired by manifold-based analyses in generative modeling~\cite{sajjadi2018assessing, kynkaanniemi2019improved, alaa2022faithful}. In contrast to methods that estimate density directly (e.g., kernel density estimation) or rely on global clustering (e.g., k-means~\cite{sajjadi2018assessing}), we use a local, non-parametric view that is better suited to high-dimensional latent spaces.
We treat each latent pose $\phi \in \Phi$ as a point on the pose manifold and approximate its local support by a $k$-NN ball under cosine distance. Specifically, we define the cosine similarity style distance by:
\begin{equation}
    d(\phi,\psi)
    = 1 - \frac{\phi^\top \psi}{\|\phi\|_2\,\|\psi\|_2},
    \label{eq:cosine_distance}
\end{equation}
and let $\mathrm{NN}_k(\phi,\Phi)$ denote the $k$-th nearest neighbor of $\phi$ in $\Phi$ under $d$. The local radius around $\phi$ is
\begin{equation}
    r_{\phi}^{(k)} = d\bigl(\phi,\mathrm{NN}_k(\phi,\Phi)\bigr),
    \label{eq:knn_radius}
\end{equation}
which we interpret as the size of the region over which a downstream model can reliably interpolate from $\phi$. \revision{\revtag{(A.2)}
Intuitively, deep models obey a local smoothness assumption: they interpolate well within regions that are sufficiently dense in the training data, but can fail abruptly once test samples lie outside this region. The radius $r_{\phi}^{(k)}$ operationalizes this idea as a data-driven ``interpolation zone'' around each pose:
a smaller $k$ corresponds to a conservative generalizer that only handles poses very close to training examples, whereas a larger $k$ corresponds to a stronger generalizer that can tolerate greater deviations. Rather than fixing a single value of $k$, we evaluate coverage across a range of $k$ values. This multi-scale view exposes both fine-grained gaps (small $k$) and broader structural deficiencies (large $k$). A 2-D illustration of the $k$-NN neighborhood construction is shown in \figref{fig:knn_coverage}: when $k = 3$, for each sample, we find its two closest neighbors (excluding itself) from the same dataset (i.e., same color) as its support.}

\subsection{Cross-Dataset Analysis: Identifying Coverage Gaps}

\begin{figure}[t]
    \centering
    \includegraphics[width=\linewidth]{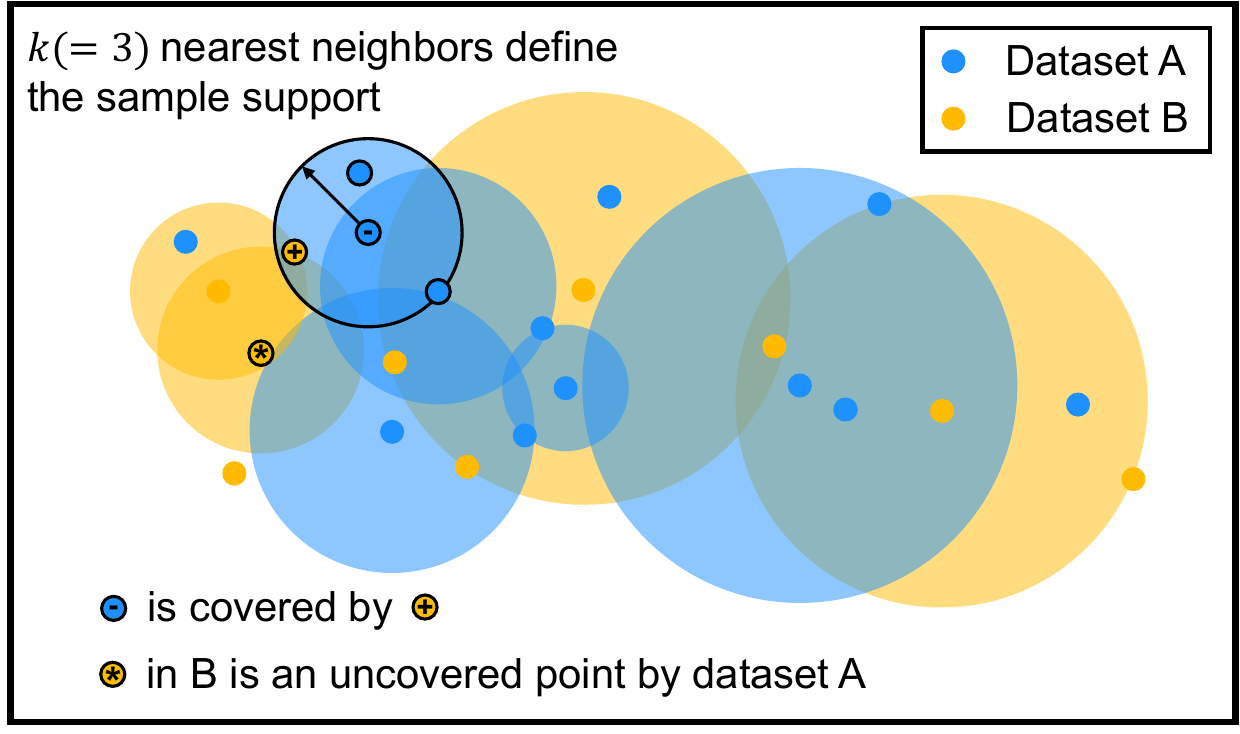}
    \caption{$k$-NN Coverage in the Latent Space.}
    \vspace{-3mm}
    \label{fig:knn_coverage}
\end{figure}

Given this notion of a local interpolation region, we define a directional \textit{Coverage} metric that answers the question:
\textit{what fraction of poses in a reference set $\Phi_A$ are close to at least one pose in a target set $\Phi_B$, relative to $\Phi_A$’s own notion of locality?}

Formally, a point $a_i \in \Phi_A$ is said to be covered by $\Phi_B$ if its nearest neighbor in $\Phi_B$ lies within its local radius $r_{a_i}^{(k)}$:
\begin{equation}
    \text{covered}(a_i; \Phi_B)
    = \mathbb{I}\Bigl( d\bigl(a_i, \text{NN}_1(a_i, \Phi_B)\bigr) \le r_{a_i}^{(k)} \Bigr),
\end{equation}
where $\mathbb{I}(\cdot)$ is the indicator function and $\text{NN}_1(a_i,\Phi_B)$ is the nearest neighbor of $a_i$ in $\Phi_B$ under $d$. The directional coverage score is then
\begin{equation}
    \text{C}(\Phi_A \to \Phi_B)
    = \frac{1}{|\Phi_A|} \sum_{a_i \in \Phi_A}
      \mathbb{I}\Bigl( d\bigl(a_i, \text{NN}_1(a_i, \Phi_B)\bigr) \le r_{a_i}^{(k)} \Bigr).
    \label{eq:coverage}
\end{equation}

As illustrated in \figref{fig:knn_coverage}, a blue point from dataset A is covered if there exists a yellow point from B within its $k$-NN support region, whereas isolated yellow points outside A’s support remain uncovered. A central byproduct of this analysis is the \textit{gap set}, $\Phi_{\text{G}}$: the subset of poses in the reference dataset that are not covered by the target dataset. When $\Phi_A$ is the large MoCap corpus and $\Phi_B$ is a mmWave dataset, $\Phi_{\text{G}}$ precisely pinpoints motions that are underrepresented in mmWave:

\begin{equation}
\label{eq:gap_set}
\Phi_{\text{G}} 
= \left\{ a_i \in \Phi_{\text{MoCap}} \,\middle|\,
d\bigl(a_i, \text{NN}_1(a_i, \Phi_{\text{mmWave}})\bigr) > r_{a_i}^{(k)} \right\}.
\end{equation}

In essence, a pose $a_i$ from MoCap is flagged as a gap if we cannot find any pose in the mmWave dataset that is close enough to it, where close enough is adaptively determined by the local density of the motion manifold around $a_i$ itself. This yields a precise and actionable candidate pool $\Phi_{\text{G}}$ for targeted data collection in \sys.

\subsection{Intra-Dataset Analysis: Evaluating Redundancy}
Next, we employ the $k$-NN coverage framework to analyze the internal structure of a single dataset. The core insight comes from the relationship between local data density, which is inversely proportional to the $k$-NN radius $r_\phi^{(k)}$, and information content:
\begin{itemize}[leftmargin=8pt]
    \item Samples in dense regions (small radii) are highly likely to be redundant. Adding a new sample that is close to many existing ones provides little new information and leads to the inefficiency we observed in our pilot study.
    \item Samples in sparse regions (large radii) are highly likely to be informative. They represent novel edge cases and are critical for improving a model's generalization.
\end{itemize}
While this principle is clear, using the raw k-NN radius $r_{\phi}^{(k)}$ directly as a comparable metric for redundancy is problematic, as its value is dependent on the dataset size $n$ and the choice of $k$. To create a more principled and scale-free measure, we introduce the \textit{Normalized Redundancy Index (NRI)}. The NRI converts the local radius into an estimate of the local ``volume'' required to contain $k$ neighbors, thereby providing a scale-free density indicator~\cite{loftsgaarden1965nonparametric,biau2015lectures}. For each sample $\phi$ in a dataset of size $n$, we define its NRI as:
\begin{equation}
\label{eq:nri}
\mathrm{NRI}_\phi(k) \;=\; \frac{(n-1)\,[r_{\phi}^{(k)}]^{\,d_\mathrm{eff}}}{k},
\end{equation}
\revision{\revtag{(B.4)}
where $d_\mathrm{eff}$ is the effective intrinsic dimension of the latent space, estimated from the slope of an regression between $\log r^{(k)}$ and $\log k$ over multiple $k$, which follows from $r^{(k)} \propto (k/n)^{1/d}$ (Refer to Appendix~\ref{app:nri} for the details). It allows us to estimate the local volume of the $k$-NN hypersphere correctly.} A smaller NRI value signifies a higher local density and thus higher data redundancy. By analyzing the distribution of NRI values across a dataset, we can rigorously quantify its overall efficiency and identify over-sampled regions.

In summary, our framework provides two key outputs: (1) A concrete gap set ($\Phi_{\text{G}}$) of missing poses, identified through cross-dataset analysis. (2) A principled NRI distribution to evaluate the internal efficiency of any dataset. The next section will focus on using this gap set to guide the high efficient data collection strategy.

\begin{figure*}[t]
    \centering
    \includegraphics[width=0.85\linewidth]{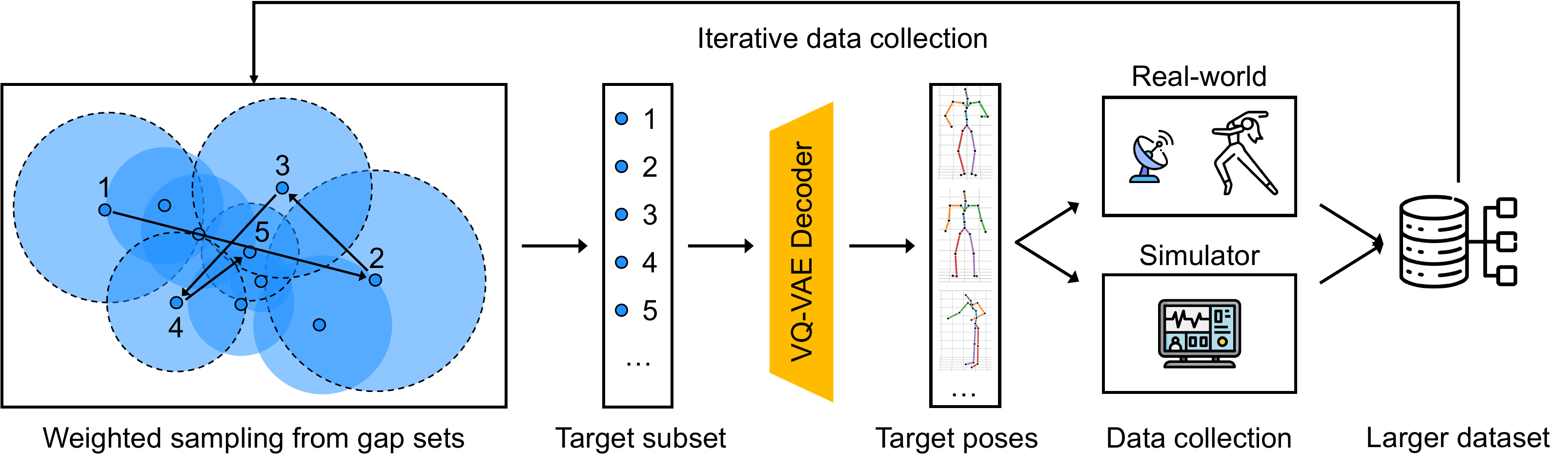}
    \vspace{-2mm}
    \caption{Coverage-driven Data Collection Workflow.}
    \vspace{-2mm}
    \label{fig:data_collection_pipeline}
\end{figure*}

\section{Coverage-Driven Data Collection}
\label{sec:data_collection}

The dataset analysis in \secref{sec:knn_framework} produces a \textit{gap set} $\Phi_G$: a large pool of candidate poses from the MoCap manifold that are insufficiently covered by the mmWave dataset. However, $\Phi_G$ is both high-volume and internally redundant. Blindly collecting all candidates would be inefficient and counterproductive. To address this, we design a coverage-driven data collection framework that is both \textit{scalable} and \textit{principled}.

As illustrated in \figref{fig:data_collection_pipeline}, our framework supports two usage modes—\textit{bootstrapping} a new dataset from scratch or \textit{augmenting} an existing one—and two acquisition pathways: \textit{real-world capture} or \textit{simulation based synthesis}. This generality ensures that the method is deployable across diverse experimental contexts.

\subsection{Target Selection via Capped-PPS Sampling}
\label{sec:cappedpps}

Given a collection budget of $B$ samples, the key task is to select a target subset $\mathcal{T} \subset \Phi_G$ that maximizes information gain while avoiding redundancy and outliers. We adopt a \textit{Capped Probability-Proportional-to-Size (Capped-PPS)} sampling strategy, which leverages local density information while applying outlier control. Specifically, for each sample $\phi_i$ in $\Phi_G$, we compute its $k$-NN radius $r^{(k)}_i$ as a proxy for local sparsity: samples with larger radii reside in underrepresented regions of the pose manifold. These radii are then converted into sampling weights:
\begin{equation}\label{eqn:sampling_weights}
    w_i = \min\left(r^{(k)}_i, \mathrm{Q}_{TH}(\{r^{(k)}\})\right)
\end{equation}
\revision{\revtag{(B.4)}
Here, $TH$ is a percentile threshold and $\mathrm{Q}_{TH}(\{r^{(k)}\})$ is the corresponding quantile value, which that acts as a validity filter. In latent space analysis, extreme distance outliers often correspond to \textit{physically implausible poses} or artifacts (e.g., severe self-intersections or floating limbs) rather than valid rare motions. By capping the weights at the $TH$-th percentile, we prevent a few extreme outliers from dominating the sampling budget and focus it on diverse yet plausible motions, rather than on implausible noise.}~\footnote{In our implementation, we set $TH$= 0.95 to filter out the last 5\% long-tail outliers and $k$= 4 based on the effective intrinsic dimension of the pose latent space calculated in~\secref{app:nri}.}. Sampling is performed using the Efraimidis--Spirakis algorithm~\cite{efraimidis2006weighted} for weighted sampling without replacement, where each candidate is assigned a priority score:
\begin{equation}
   p_i = -\log u_i / w_i,\quad u_i \sim \mathcal{U}(0,1)
\end{equation}

This method prioritizes sparse regions with higher sampling probability for diversity and avoids budget waste on implausible or outlier poses for robustness. It supports both data augmentation and bootstrapping. In augmentation mode, the candidate space is the known $\Phi_G$; in bootstrapping mode, the candidate space is the full pose space.

\subsection{Data Acquisition Pipeline}
\label{sec:pipeline}

Once the $B$ target embeddings $\mathcal{T}$ are selected, we instantiate them as samples using a flexible acquisition pipeline:

\begin{itemize}[leftmargin=8pt]
    \item \textbf{Pose Reconstruction.} For each latent sample $\phi \in \mathcal{T}$, the trained VQ-VAE decoder reconstructs a clean, canonical 3D skeleton $\mathbf{x}$. This ensures all downstream data generation is grounded in a unified latent representation.
    
    \item \textbf{Pose Realization and Data Capture.} The reconstructed skeleton is then realized for data capture in one of two modes. For real-world capture, the skeleton is rendered visually and mimicked by a human participant for radar capture. For synthetic capture, the skeleton is converted to an SMPL-X mesh and imported into a mmWave sensor simulator, which computes realistic radar returns based on pose, position, and radar parameters.

    \item \textbf{Stopping Criterion.} \revision{\revtag{(D.2)} Unlike conventional pipelines with ambiguous termination points, \sys defines a clear stopping condition via the gap set $\Phi_G$. As $\Phi_G$ represents the finite set of uncovered motions, collection concludes when $\Phi_G$ falls below a density threshold. In practice, we operate under a fixed budget $B$, strictly prioritizing the highest-weighted targets in $\Phi_G$ to maximize coverage.}
\end{itemize}

After each acquisition round, the newly collected data is added to $\Phi_{\text{mmWave}}$, and the coverage analysis in \secref{sec:knn_framework} is recomputed. This enables an \textit{iterative closed-loop acquisition process}: each round targets new coverage gaps, steadily improving dataset diversity, efficiency, and generalization power.
\section{Evaluation}
\label{sec:evaluation}
Our evaluation is designed to provide a systematic and evidence-based validation of the proposed framework. To this end, we first report the end-to-end performance of our data collection method. Then we conduct a series of microbenchmarks and measurement studies to examine the validity of our latent space representation and metric reasonableness.

\subsection{Experimental Setup}
\label{sec:setup}

To ensure rigor and reproducibility, we design our experiments around a combination of large-scale motion capture priors, representative mmWave datasets, a standardized training environment, simulation pipeline, benchmark dataset, and real-world experiment setups.

\nosection{Vision MoCap Datasets}
We adopt the AMASS dataset~\cite{mahmood2019amass} as our comprehensive motion capture reference. AMASS aggregates dozens of optical HPE datasets into a unified SMPL-X parameterization and contains over 8.9M motion sequences. 
Its scale and diversity make it a suitable surrogate for the broader human pose manifold, serving as the reference prior for our coverage analysis.

\nosection{MmWave HPE Datasets}
Our evaluation focuses on two widely used mmWave HPE datasets: mmBody~\cite{chen2022mmbody} and MMFi~\cite{mmfi}. We choose them because they are the largest and most representative publicly available datasets with reliable annotations, while other resources either suffer from alignment and quality issues or adopt bespoke formats incompatible with our pipeline. 
mmBody provides SMPL-X parameterized labels of 22 keypoints human skeletons, whereas MMFi follows the Human3.6M skeleton (17 keypoints) with additional subject and action annotations. Since our VQ-VAE operates on SMPL-X inputs, we convert MMFi labels via a MoCap-to-SMPL-X pipeline~\cite{pavlakos2019expressive, Lv_2024_CVPR} to ensure consistency.

\nosection{Model and Training Environment}
Our universal pose tokenizer is a VQ-VAE with an MLP-Mixer backbone~\cite{geng2023human}, trained exclusively on AMASS. The human pose estimator is a Point-Transformer~\cite{zhao2021point} model identical to our pilot study (\secref{sec:motivation}). All $k$-NN operations for coverage analysis are accelerated with FAISS-GPU~\cite{johnson2017billion}. Experiments are run on a server equipped with a Genuine Intel(R) CPU, 32 GB RAM, and six NVIDIA GeForce RTX 3090 GPUs (24 GB each). We use a batch size of 16 and train for 25 epochs.

\nosection{Simulation Pipeline} 
We evaluate the data scaling efficiency under simulation environments to avoid domain gap and ensure fairness among different methods. We adapt RF Genesis~\cite{chen2023rf}, a ray-tracing-based simulator for mmWave HPE. The simulated radar is configured with a 77 GHz carrier frequency, 3.5 GHz bandwidth, a 3Tx–4Rx antenna array, and 256 ADC samples. It operates at 10 frames per second, with each frame comprising 128 chirps. The original simulator takes textual pose descriptions, generates SMPL-X meshes via a diffusion model~\cite{tevet2022human}, and outputs radar range Doppler spectra and point clouds. We modify its pipeline to accept single-frame skeleton inputs directly, bypassing text-to-pose synthesis. We manually validate the generated point clouds for geometric plausibility and cross-verify them against reference SMPL-X poses. To evaluate out-of-distribution (OOD) generalization, we construct a challenging benchmark of 40k poses uniformly sampled from the latent pose space of $\Phi_{\text{AMASS}}$, which represents all possible and reasonable human poses. This benchmark is strictly held out from all training and validation, ensuring no data leakage.

\nosection{Real-World Data Collection Pipeline}
We also build a real-world acquisition pipeline. Our mmWave front-end is a 3Tx–4Rx FMCW radar~\cite{duopucetan2025} operating at 79\,GHz with 6\,GHz bandwidth and 15 frames per second, placed about 3 meters in front of the subject. 
It is paired with an Intel RealSense D455 RGB-D camera and a four-camera vision motion capture system~\cite{shuai2022multinb, easymocap} that provides 3D ground-truth poses. The cameras are rigidly mounted around the sensing area, and we perform joint spatial and temporal calibration to align radar, RGB-D, and MoCap coordinate frames, ensuring accurate human-pose–mmWave point-cloud correspondence. The RGB-D stream is used only for calibration; supervision comes solely from the vision MoCap labels.

\subsection{Data Scaling Efficiency} 
\label{ssec:simulation_exp}

We ask whether our coverage-aware acquisition policy yields better OOD scaling than the conventional sequential-action protocol under the same data budget. To answer this, we evaluate the data scaling efficiency of \sys with simulation data. We compare two acquisition strategies, each generating up to 40k samples (matching the size of the mmBody training set) and evaluated on our OOD generalization benchmark:  
(i) our proposed \sys, following the sampling weights in \eqnref{eqn:sampling_weights}, and  
(ii) a baseline designed to mirror existing mmWave collection practices.

Since we cannot replicate the hardware and experimental conditions of existing open sourced datasets, all comparisons are conducted in a controlled simulation environment. Specifically, we reproduce the mmBody dataset’s sampling trajectory by feeding its ground-truth skeleton labels into the simulation pipeline, producing a ``simulated mmBody'' dataset. This baseline reflects the prevailing methodology in existing datasets: predefined actions are captured sequentially without regard to motion diversity or distributional coverage. Each strategy yields a separate synthetic dataset, on which we train identical models using the same configuration; we then measure OOD MPJPE on the held-out generalization benchmark.

\begin{figure}[t]
    \centering
    \includegraphics[width=\linewidth]{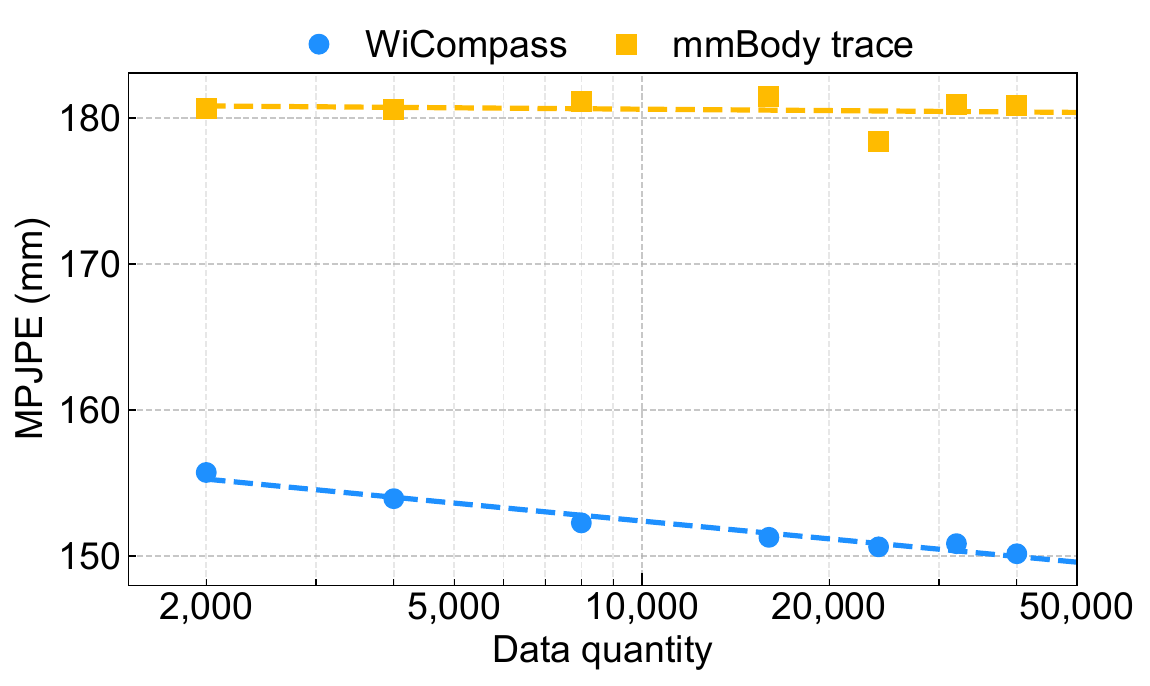}
    \vspace{-8mm}
    \caption{Scaling Laws Comparison on Synthetic Data.}
    \vspace{-4mm}
    \label{fig:scaling_laws_simulation}
\end{figure}

\begin{figure*}[t]
    \centering
    
    \begin{subfigure}{0.32\linewidth}
        \centering
        \includegraphics[width=\linewidth]
        {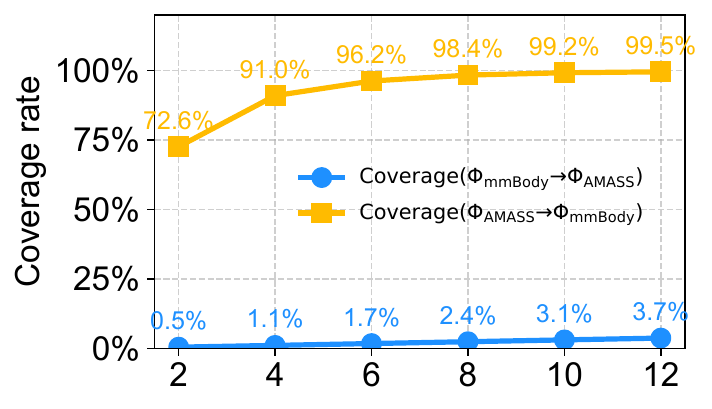}
        \vspace{-7mm}
        \subcaption{mmBody $\leftrightarrow$ AMASS.}
        \label{fig:coverage_mmbody}
    \end{subfigure}
    \hfill
    \begin{subfigure}{0.32\linewidth}
        \centering
        \includegraphics[width=\linewidth]{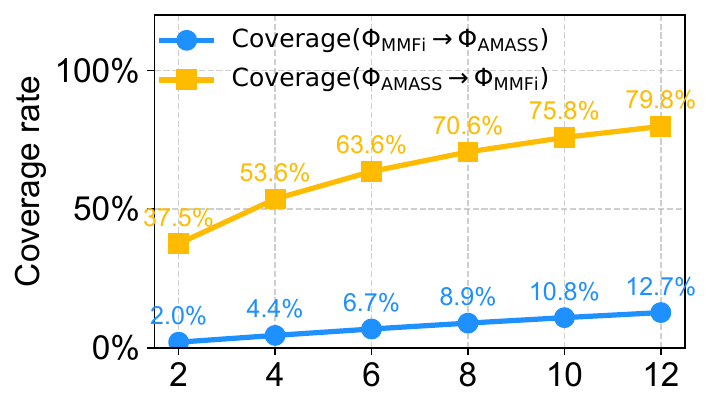}
        \vspace{-7mm}
        \subcaption{MMFi $\leftrightarrow$ AMASS.}
        \label{fig:coverage_mmfi}
    \end{subfigure}
    \hfill
    \begin{subfigure}{0.32\linewidth}
        \centering
        \includegraphics[width=\linewidth]{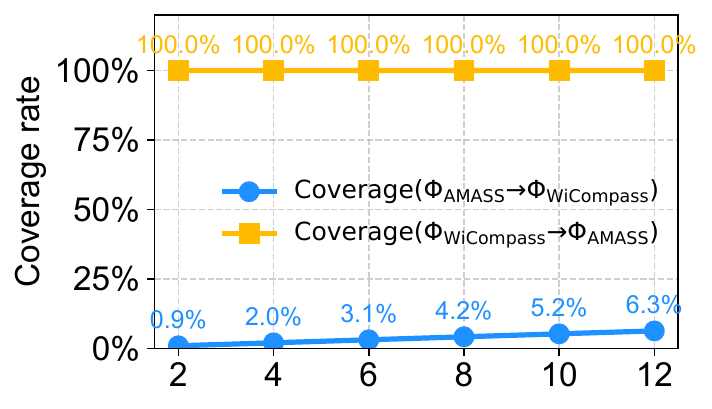}
        \vspace{-7mm}
        \subcaption{\sys $\leftrightarrow$ AMASS.}
        \label{fig:coverage_wicompass}
    \end{subfigure}
    \vspace{-3mm}
    \caption{Cross-Dataset Coverage vs. $k$.}
    \vspace{-2mm}
    \label{fig:cross_coverage}
\end{figure*}

\begin{figure*}[t]
    \centering
    \begin{subfigure}[t]{0.24\linewidth}
        \centering
        \includegraphics[width=\linewidth]{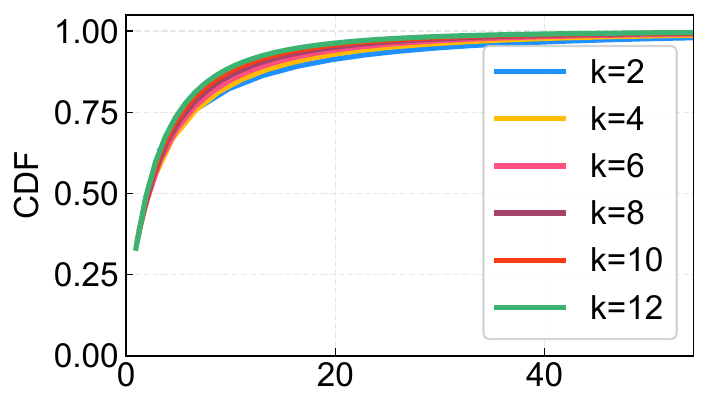}
        \vspace{-5mm}
        \caption{AMASS.}
        \label{fig:cdf_nri_amass}
    \end{subfigure}
    \hfill
    \begin{subfigure}[t]{0.24\linewidth}
        \centering
        \includegraphics[width=\linewidth]{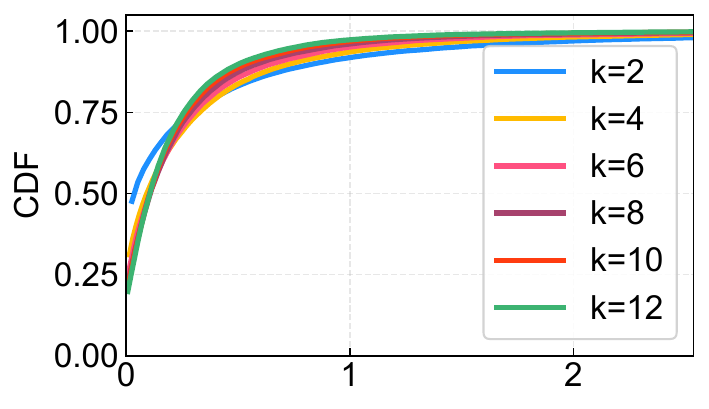}
        \vspace{-5mm}
        \caption{mmBody.}
        \label{fig:cdf_nri_mmbody}
    \end{subfigure}
    \hfill
    \begin{subfigure}[t]{0.24\linewidth}
        \centering
        \includegraphics[width=\linewidth]{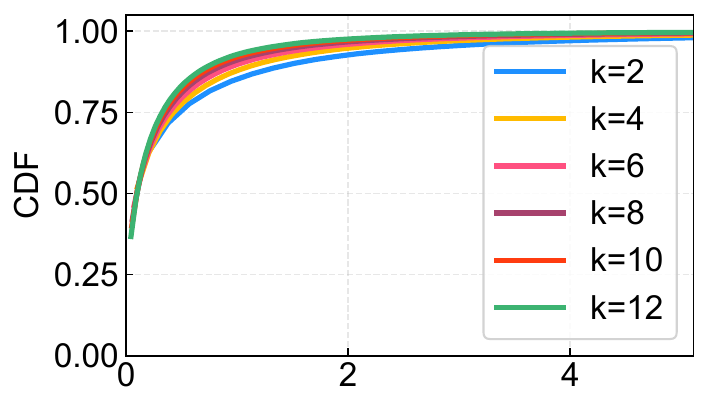}
        \vspace{-5mm}
        \caption{MMFi.}
        \label{fig:cdf_nri_mmfi}
    \end{subfigure}
    \hfill
    \begin{subfigure}[t]{0.24\linewidth}
        \centering
        \includegraphics[width=\linewidth]{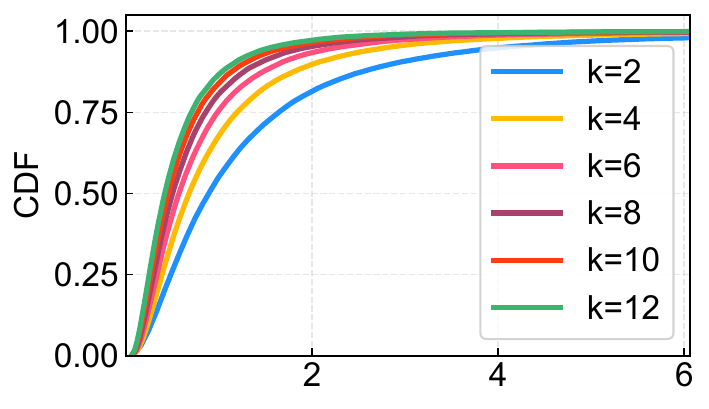}
        \vspace{-5mm}
        \caption{\sys.}
        \label{fig:cdf_nri_wicompass}
    \end{subfigure}
    \vspace{-3mm}
    \caption{NRI Distribution for Intra-Dataset Coverage Analysis.}
    \vspace{-2mm}
    \label{fig:nri}
\end{figure*}

\nosection{Results}
To quantify how performance evolves with scale, we adopt a power-law model for data scaling~\cite{kaplan2020scaling}:
\begin{equation}
   E(D) = A D^{\alpha_D},
\end{equation}
where $D$ is the dataset size, $E(D)$ is the testing error, $A$ is a fitted constant, and $\alpha_D$ is the data-scaling exponent that characterizes how rapidly performance changes with additional data samples. We fit this model via nonlinear least-squares regression on log–log transformed data, and the resulting curves in \figref{fig:scaling_laws_simulation} reveal a clear difference between coverage-aware and conventional collection.
For \sys, the best-fit curve is $ E_{\sys}(D) \approx 169.57\,D^{-0.0116} $, with a strong log–log correlation ($r \approx -0.98$), indicating a stable power-law regime. In contrast, the mmBody-trace baseline is fit by $E_{\text{mmBody-trace}}(D) \approx 181.86\,D^{-0.0008}$, with a much weaker correlation ($r \approx -0.15$).
As we increase the simulated dataset size, \sys yields a consistent monotonic decrease in OOD MPJPE, with an order-of-magnitude larger (more negative) scaling exponent than the baseline. This indicates that additional samples collected under the coverage-aware policy remain informative even at tens of thousands of frames. Using these fits, \sys reduces OOD MPJPE by roughly 25–30\,mm compared to the baseline across 2k–40k samples, and the gap widens as $D$ grows.

\subsection{Dataset Coverage Measurement}\label{ssec:coverage_evaluation}
We next relate the scaling behaviors to coverage metrics in the latent pose space. We use the $k$-NN coverage metric (\secref{sec:knn_framework}) to evaluate the pose coverage of existing mmWave datasets (mmBody, MMFi) and our collected dataset \sys against the AMASS reference, and to quantify their internal redundancy.

\nosection{Cross-Dataset Coverage}
Coverage is defined as the fraction of MoCap poses that are supported—i.e., have at least one mmWave sample within their local neighborhood of radius $r^{(k)}$. Fig.~\ref{fig:cross_coverage} shows directional coverage as a function of $k$. Key observations include:


\begin{itemize}[leftmargin=10pt]
    \item Coverage increases smoothly with $k$ for all datasets, indicating that our metric behaves consistently.
    
    \item mmBody covers a narrow subregion of AMASS. In \figref{fig:coverage_mmbody}, coverage of AMASS by mmBody (blue) reaches only 3.7\% at $k=12$, while the reverse direction (yellow) exceeds 99\%. This suggests that mmBody occupies a narrow, but valid, subset of the MoCap manifold.

    \item MMFi exhibits limited and noisy coverage. In \figref{fig:coverage_mmfi}, coverage of MMFi by AMASS (yellow) reaches 79.8\% at $k=12$, below mmBody. We attribute this to label noise introduced during Human3.6M-based annotation—specifically, missing joints (e.g., neck, clavicles, ankles) cause alignment errors despite the use of optimization-based format conversion. Nevertheless, the reverse coverage (blue) still reveals distinctive motion regions that MMFi fails to capture, highlighting opportunities for future data acquisition.

    \item \sys achieves superior coverage efficiency. In \figref{fig:coverage_wicompass}, coverage of \sys by AMASS (yellow) is 100\% since \sys is generated from the AMASS pose space. More importantly, coverage of AMASS by \sys (blue) consistently exceeds that of mmBody and comes within a factor of two of MMFi, while using roughly one-eighth as many samples. This indicates that our sampling method achieves broader and more representative coverage per unit of data budget.

\end{itemize}

\nosection{Intra-Dataset Coverage}
To assess internal redundancy, we compute the cumulative distribution function (CDF) of NRI, as shown in \figref{fig:nri}. For each dataset, we set the x-axis limits from $\min_k Q_{0}(\mathrm{NRI}_k)$ to $\max_k Q_{98}(\mathrm{NRI}_k)$ across all plotted $k$ values to suppress extreme outliers while keeping all CDF curves visible. Key observations include:
\begin{itemize}[leftmargin=10pt]
    \item AMASS shows broad coverage with informative samples. \figref{fig:cdf_nri_amass} reveals a wide NRI spread, reflecting diverse and sparsely sampled poses. It serves as a reference baseline for well-structured motion coverage.

    \item MmBody (\figref{fig:cdf_nri_mmbody}) suffers from high redundancy. Its CDF rises steeply with uniformly low NRI values, confirming that mmBody contains many highly redundant samples. MMFi (\figref{fig:cdf_nri_mmfi}) exhibits moderate redundancy.
    
    \item \sys encourages sparse, informative sampling. \figref{fig:cdf_nri_wicompass} exhibits the broadest NRI distribution among all mmWave datasets, validating that our coverage-driven acquisition strategy effectively targets underrepresented regions of the motion manifold.
\end{itemize}
The consistency of NRI curves across different neighborhood sizes $k$ further supports its utility as a \textit{robust and scale-insensitive metric} for evaluating intra-dataset redundancy.

\nosection{Metrics as Generalization Proxies}
We further validate that these metrics serve as robust, predictive proxies for generalization—rather than post-hoc descriptors—through two steps: independent diagnosis and predictive confirmation. First, the metrics correctly diagnose the failures observed in our pilot study on existing datasets (\secref{ssec:bottleneck}). For instance, the poor OOD generalization aligns with our finding that mmBody covers only 3.7\% of the pose manifold (\figref{fig:coverage_mmbody}). Likewise, the finding that 70\% of data is discardable aligns with our NRI analysis, which exposes massive structural redundancy (\figref{fig:cdf_nri_mmbody}). This confirms the metrics are valid independent of \sys. Second, the metrics accurately predict \sys's efficiency gains. By targeting the identified coverage gaps, \sys achieves both broader metric distribution and significantly lower OOD error (\secref{ssec:simulation_exp}). This direct correlation between metric improvement and error reduction confirms that $k$-NN coverage and NRI are predictive proxies that effectively guide data quality.

\subsection{Real-world Validation}




\begin{figure}[t]
    \centering

    \makebox[\linewidth][c]{%
        \begin{subfigure}[t]{0.42\linewidth}
            \centering
            \includegraphics[width=\linewidth]{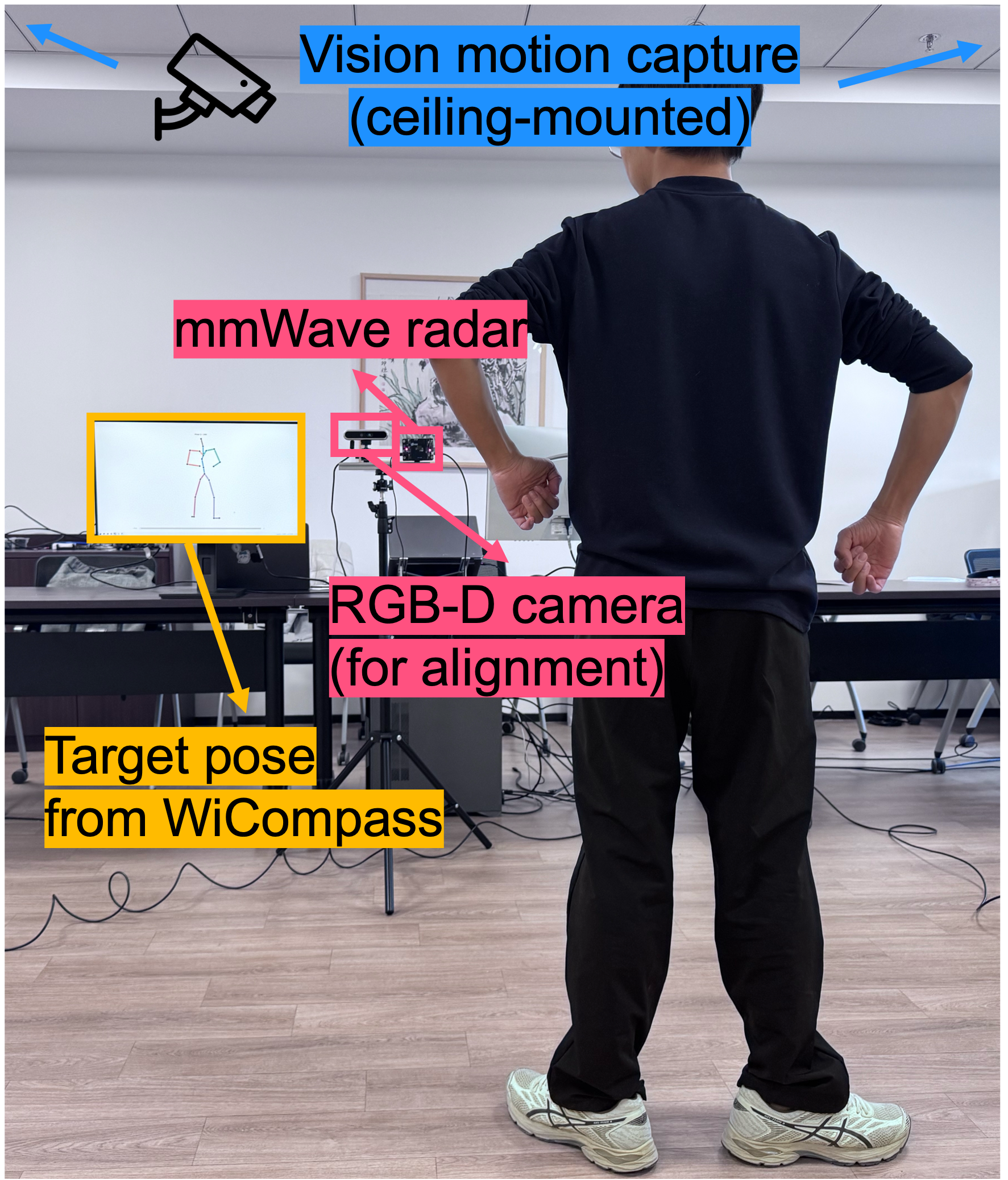}
            \vspace{-4mm}
            \subcaption{Setups.}
            \label{fig:realworld_exp_setup}
        \end{subfigure}\hspace{0.035\linewidth}%
        \begin{subfigure}[t]{0.545\linewidth}
            \centering
            \includegraphics[width=\linewidth]{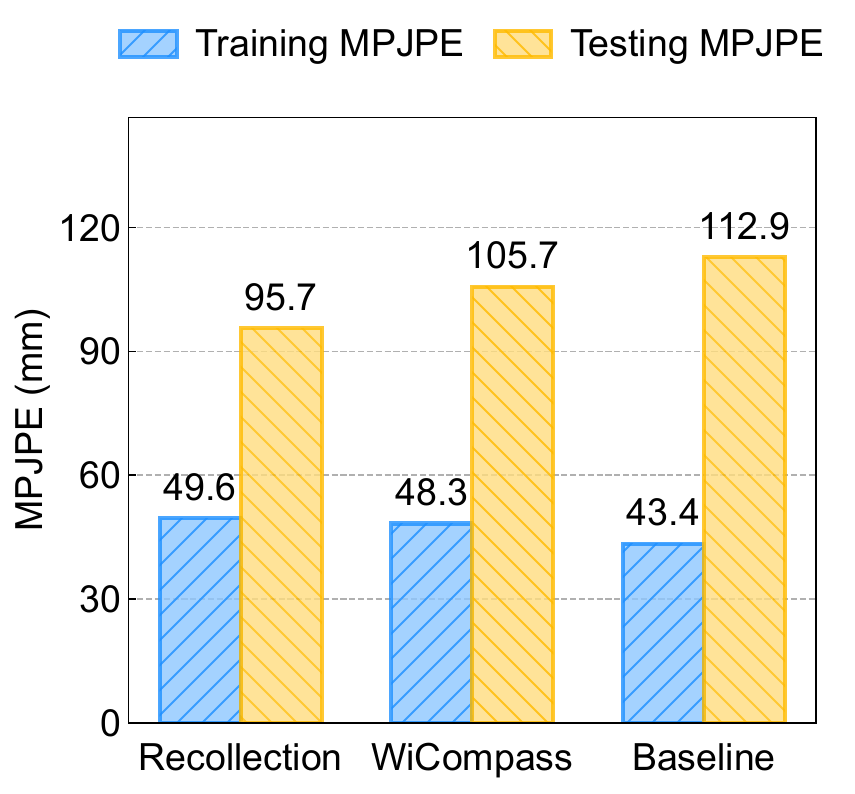}
            \vspace{-4.2mm}
            \subcaption{Generalization Errors.}
            \label{fig:realworld_results}
        \end{subfigure}%
    }

    \vspace{-2mm}
    \caption{Real-world Validation.}
    \label{fig:realworld_validation}
    \vspace{-5mm}
\end{figure}

We conduct a closed-loop real-world feasibility validation, demonstrating that \sys can instantiate concrete acquisition targets and drive practical mmWave HPE data collection that improves generalization under a fixed budget.

\nosection{Protocol and Acquisition Strategies}
We choose bodyweight exercises as a challenging case study because its large-amplitude, whole-body motions are substantially more diverse than the upright postures dominating existing mmWave datasets. Two volunteers performed exercise sequences to form a held-out benchmark of 8k frames, which is strictly reserved for testing. Under an identical budget of 8k training frames, we compare three acquisition strategies:
(i) Recollection. The same motion family (bodyweight exercises) is captured again, providing the closest possible distributional match to the benchmark;
(ii) \sys. We project the target motion family into the universal latent pose space and use our acquisition policy to select informative target poses; the volunteer then mimics these on-screen visualizations for radar capture;
(iii) baseline. Following common practice in prior datasets (e.g., MMFi), we collect data by sequentially recording 10 common actions (i.e., standing, walking, jogging, turning, squatting down and up, waving left/right, raising left/right hand, and jumping).

\nosection{Results} 
\figref{fig:realworld_validation} reports MPJPE on the training split and on the held-out benchmark. All methods achieve comparable training errors (43.4--49.6\,mm), but their generalization differs substantially. The recollection set attains the lowest test error (95.7\,mm), as expected: by construction, its acquisition list exactly matches the poses in the benchmark, so the training and test distributions are identical. It therefore serves as an oracle upper bound rather than a realistic data-collection strategy. In this context, \sys achieves a test MPJPE of 105.7\,mm, moving significantly closer to the recollection oracle, and clearly outperforming the baseline (112.9\,mm). Notably, the baseline attains the lowest training error (43.4\,mm) yet the worst test error, underscoring that conventional action-list collection can be sample-inefficient for generalization. Although absolute errors remain relatively high due to the limited real-world scale (8k frames), this feasibility study demonstrates an end-to-end, practical closed-loop pipeline and provides direct evidence that \sys prioritizes more informative real mmWave samples under a fixed collection budget.

\subsection{VQ-VAE Microbenchmarks}\label{ssec:framework_evaluation}
The latent space extracted from the VQ-VAE bottleneck layer is the foundation of our framework. We evaluate whether the learned pose vocabulary is compact, representative, and semantically interpretable.

\begin{figure}[t]
    \centering
    \includegraphics[width=0.9\linewidth]{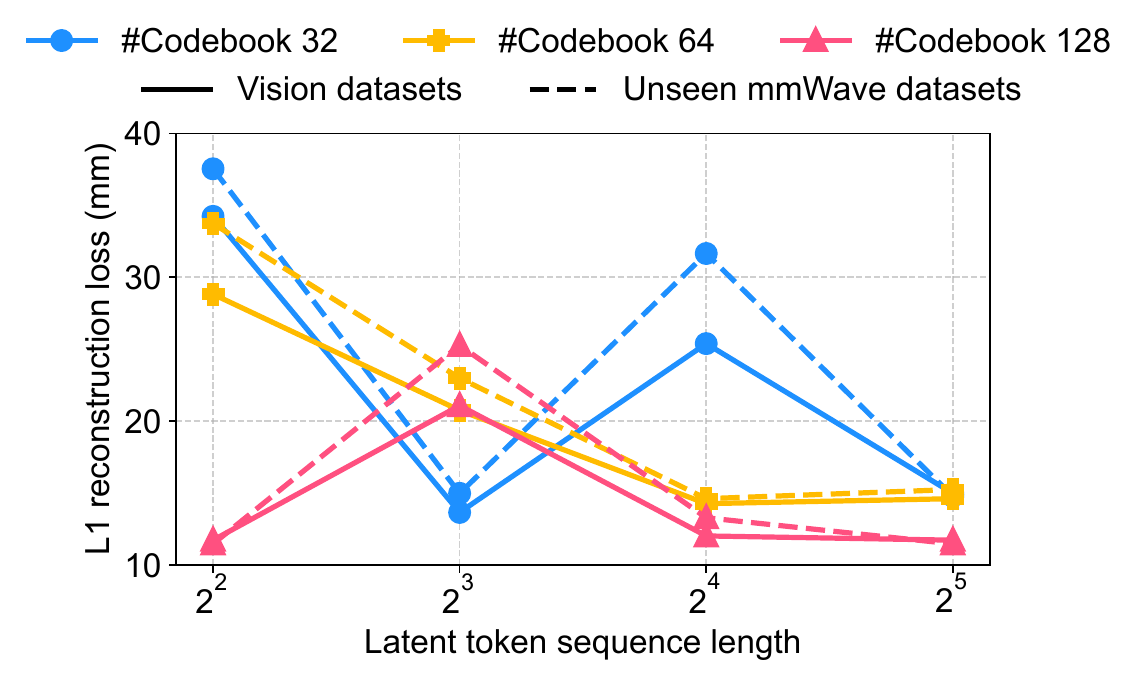}
    \vspace{-3mm}
    \caption{Reconstruction Fidelity of VQ-VAE Model.}
    \vspace{-2mm}
    \label{fig:vqvae_performance}
\end{figure}

\begin{figure}[t]
    \centering
    \includegraphics[width=\linewidth]{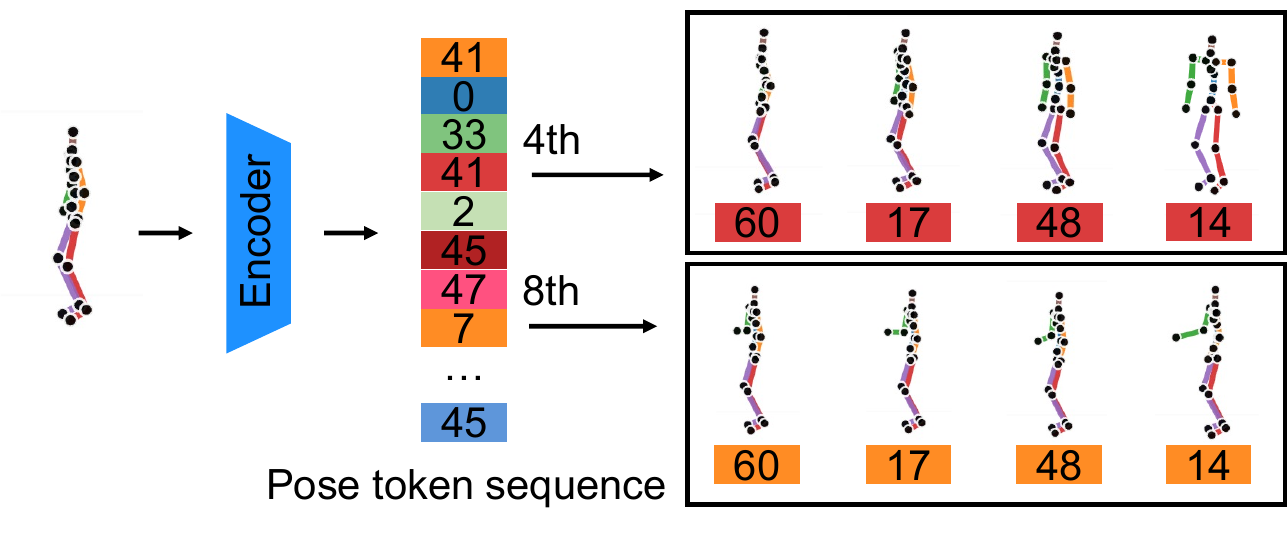}
    \vspace{-6mm}
    \caption{Pose Token Influence Visualization.}
    \vspace{-4mm}
    \label{fig:token_vis}
\end{figure}

\nosection{Reconstruction Fidelity of VQ-VAE Model}
The goal of the VQ-VAE model is to encode human poses into a discrete latent space that is both semantically meaningful and computationally efficient. Two hyperparameters are critical: the codebook size $K$ and the latent sequence length $T$. Smaller $K$ and $T$ maximize compression efficiency and enforce semantic compactness, but overly small values risk under-representing the pose manifold and causing high reconstruction error. We compute the L1 reconstruction loss on two datasets: the AMASS test set and the unseen mmBody dataset. ~\figref{fig:vqvae_performance} plots the average reconstruction error across different latent configurations, varying the codebook size and sequence length. 
We make the following remarks:
\begin{itemize}[leftmargin=10pt]
    \item Generally, the reconstruction fidelity decreases with larger $K$ or $T$. At $K=64$, $T=16$, the reconstruction error stabilizes at approximately 10\,mm on both AMASS and mmBody. 
    This error is on par with the groundtruth error of MoCap systems~\cite{mundermann2006evolution, tautges2011motion}. After this point, increasing $K$ or $T$ slightly improves fidelity but yields diminishing returns. Therefore, we select $K=64$, $T=16$ as the default configuration in our all experiments.
    \item These results confirm that the latent representation generalizes across modalities, supporting downstream use in coverage analysis.
\end{itemize}

\nosection{Qualitative Visualization} 
\figref{fig:token_vis} illustrates token-level influence by modifying one latent token at a time in mmBody poses. The results show that tokens correspond to distinct motion primitives—for example, the fourth token controls global orientation, while the eighth affects the right elbow. This demonstrates the semantic interpretability of the codebook, consistent with prior findings in pose-space studies~\cite{guo2022tm2t, zhang2023generating, saleem2025genhmr}. Additional visualization cases, including poses from existing mmWave HPE datasets as well as those uncovered relative to AMASS, are provided in Appendix~\ref{app:poses}.
\section{Related Work}

\nosection{MmWave HPE}
Recent mmWave HPE researches follow three major directions. 
\textit{(1) Signal representation.} Models either operate directly on raw radar cubes with 3D CNNs or Transformers, project RF signals into 2D range–angle or range–Doppler views for heatmap prediction, or reconstruct sparse point clouds for point-based backbones with temporal aggregation \cite{chen2025cpformer,kini2024transhupr,chen2022mmbody}. 
\textit{(2) Structure-aware inference.} Body priors improve robustness under sparsity and occlusion, including graph–attention hybrids for part refinement and diffusion-style denoisers that regularize kinematics \cite{lee2023hupr,mmdiff}. 
\textit{(3) Multi-view and cross-modality fusion.} Multi-view radar mitigates directionality and self-occlusion via learned fusion, while vision-derived labels bootstrap radar-only models without manual annotation \cite{choi2025mvdoppler,lee2023hupr}. 
Despite these advances, all approaches depend on existing datasets so struggle with poor generalization to unseen actions, subjects, and environments. Our work tackles this limitation from a distributional perspective, orthogonal to architectural design.

\nosection{MmWave HPE Data Scaling}\label{ssec:related_mmwave_data}
Scaling datasets and simulators has been another major driver of progress. Public corpora such as mRI, mmBody, MMFi, MilliPoint, HuPR, and RT-Pose enable supervised or cross-modality training, but their coverage remains uneven and limited across subjects, actions, and environments \cite{an2022mri,chen2022mmbody,mmfi,cui2023milipoint,lee2023hupr,ho2024rt}, as we measured in the \secref{ssec:bottleneck} and \secref{ssec:coverage_evaluation}. 
Synthetic pipelines—ranging from video-to-radar translation~\cite{ahuja2021vid2doppler,deng2023midas,zhang2022synthesized} to hybrid physics–generative methods~\cite{chen2023rf,li2024sbrf}—expand cheaply but face sim-to-real gaps (multipath, materials, hardware) and often impose distribution shift problem~\cite{gong2025data}. 
Our approach is complementary: rather than indiscriminately enlarging datasets, we provide a distribution-aware criterion that directs both real and synthetic collection toward underrepresented regions.

\nosection{Active Learning and Core-Set}\label{ssec:related_al}
Our objective of optimizing data collection parallels active learning (AL) and core-set selection. Classical AL methods optimize sample efficiency by querying the most informative points from a large unlabeled pool using uncertainty or diversity metrics (e.g., entropy, margins, query-by-committee)~\cite{lewis1995sequential,seung1992query,houlsby2011bayesian,gal2017deep}, while core-set methods emphasize geometric diversity through $k$-center or submodular objectives~\cite{sener2017active,wei2015submodularity,ash2019deep,mirzasoleiman2020coresets,killamsetty2021grad,killamsetty2021glister}. Crucially, however, these approaches are fundamentally \textit{selective}: they assume a large, pre-collected pool of candidate sensor data and operate by reweighting or filtering this pool. In contrast, \sys adopts an \textit{oracle-guided, plan-based acquisition} paradigm. Rather than scoring an existing pool, it constructs an acquisition plan directly in the MoCap oracle pose space and then executes this plan in the radar domain, synthesizing novel, informative targets that do not yet exist as mmWave measurements. This shifts the workflow from ``filtering a pool'' to ``synthesizing a plan,'' bypassing the need for massive and blind data pre-collection.

\nosection{Pose Latent Space}
Latent variable models, particularly VQ-VAEs, are established tools in computer vision for motion generation and compression. We emphasize that our contribution is not the architecture itself, but the \textit{novel repurposing} of its latent space as a diagnostic framework for wireless sensing. While prior work restricts latent spaces primarily to instance-level tasks (e.g., reconstruction or generation)~\cite{pavlakos2019expressive, guo2022tm2t, zhang2023generating, saleem2025genhmr, zhou2025page}, we utilize the discrete latent space as a \textit{distributional probe}. By projecting disparate sources—optical MoCap priors and mmWave pose labels—into this shared density manifold, we translate abstract concepts of ``dataset coverage'' and ``redundancy'' into computable metrics ($k$-NN radius and NRI). This enables rigorous quantification of mmWave dataset quality in a manner inaccessible to raw signal analysis alone.
\section{Discussion}

\nosection{Latent Space Completeness}
The utility of our latent pose space hinges on the diversity of its underlying motion data. We leverage AMASS~\cite{mahmood2019amass}, a de facto human motion prior aggregating 40 hours of diverse captures—including daily actions (CMU, KIT), object interactions (GRAB), and extreme postures (MOYO). However, completeness remains an ideal; culturally specific gestures and rare outdoor behaviors are still underrepresented, as evidenced by the coverage never achieve 100\% in \figref{fig:cross_coverage}. Furthermore, VQ-VAE discretization, while enabling tractable analysis, inevitably merges subtle variations into shared codewords, reducing sensitivity to fine-grained differences. Future work should explore hierarchical quantization to capture more nuanced motion structures.

\nosection{Sim-to-Real Gap}
Our simulator serves as an efficient testbed for evaluating the \textit{relative} sample efficiency of acquisition strategies. We acknowledge that it does not fully capture real-world complexities such as multipath propagation and hardware noise; thus, absolute performance metrics in simulation may not directly transfer to deployment. However, the selection signal in \sys relies on modality-agnostic pose labels in the shared latent space rather than raw radar features. This design ensures that our target selection policy remains invariant to the sim-to-real gap, even if the downstream radar training is domain-dependent. Furthermore, ensuring identical simulation protocols across all baselines guarantees that observed gains arise strictly from the acquisition strategy rather than data artifacts.

\nosection{Evaluation Scale and Extrapolated Conclusions}
Our experiments at the 40k-sample scale (simulation) and 8k-sample scale (real-world) demonstrate that \sys offers a superior acquisition strategy. However, verifying whether the observed log–log scaling behavior persists at the massive scales seen in vision and language models remains an open question. MmWave sensing involves fundamentally different data sparsity patterns, and we are currently constrained by the prohibitive cost of large-scale real-world collection. While our fitted curves provide meaningful extrapolations within the tested regime, we caution against generalizing these trends to arbitrarily large scales without further validation. Future work will focus on large-scale, multi-round real-world studies to explore these limits.

\nosection{Generalization Across Multiple Dimensions}
This work primarily addresses the \textit{pose diversity} (geometric coverage) to ensure the model observes the comprehensive manifold of human kinematics. We recognize that real-world robustness also demands generalization across physical environments (e.g., multipath and clutter). We view this ``sensing gap'' as distinct from the kinematic gap addressed by \sys; they are orthogonal problems that must be solved in tandem. A key strength of our framework is its extensibility: the proposed coverage analysis are domain-agnostic. Future work can extend this framework to environmental latent spaces, enabling the joint targeting of kinematic and environmental diversity for holistic wireless sensing.
\section{Conclusion}
This paper shows that the key bottleneck in mmWave HPE lies not in model/algorithm but in data. By quantifying coverage and redundancy in a modality-agnostic latent space, we transform vague concerns about diversity into actionable diagnostics and a principled acquisition strategy. Our coverage-driven loop consistently improves generalization comparing to the baseline trajectories. More broadly, the lesson for wireless sensing is clear: prioritize data quality over scale, treat datasets as engineered artifacts with measurable coverage, and guide costly RF data collection with inexpensive cross-modal priors. By shifting attention to improving data quality, \sys provides a reproducible path toward robust, data-driven wireless sensing.

\section*{Acknowledgments}
This work is supported by National Key R\&D Program of China (Grant No. 2023YFF0725004), National Natural Science Foundation of China (Grant No. U25B2040, 62502012, and 62272010), and the Postdoctoral Fellowship Program and China Postdoctoral Science Foundation (Grant No. GZC20251070). Chenren Xu is the corresponding author.

\ifeg
\else
\fi

\bibliographystyle{unsrt} 
\bibliography{ref}

\appendix
\section{Model Architecture and Scaling}
\label{app:model}

\begin{figure}[h]
    \centering
    \includegraphics[width=\linewidth]{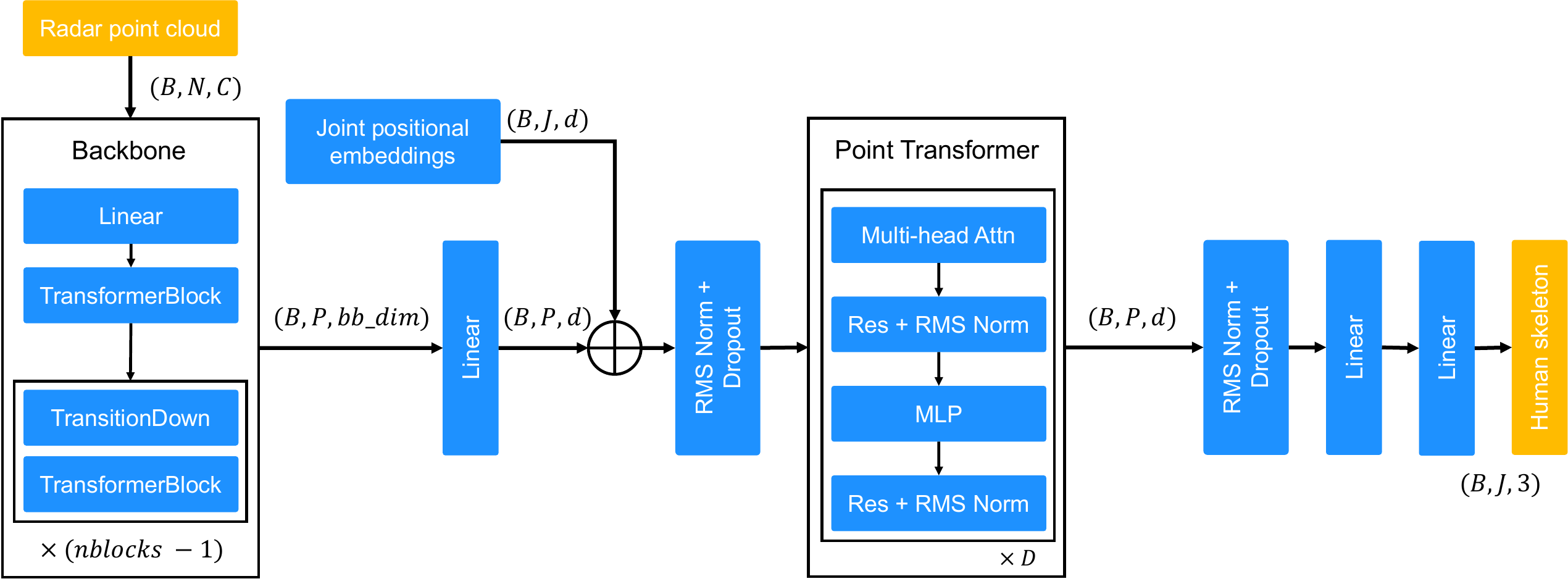}
    \vspace{-3mm}
    \caption{Our mmWave HPE Model.}
    \vspace{-4mm}
    \label{fig:model_architecture}
\end{figure}

\nosection{Model Structure}
Our model, shown in \figref{fig:model_architecture}, consists of three stages: 
(i) a lightweight backbone encodes radar points into per-point features, 
(ii) a fixed set of learnable \emph{joint tokens} are concatenated with point tokens and processed by a Transformer encoder stack, and 
(iii) two small MLP heads decode the joint tokens into compact features and final 3D coordinates.

\nosection{Scaling Strategy}
We vary model size using a \emph{depth-first compound scaling} rule: 
the primary axis is the number of Transformer layers ($D$); 
the per-head dimension is fixed at $\mathrm{dim}_{\text{head}}=64$; 
the number of heads is modestly increased so that the model width is $d=\text{heads}\times 64$; 
and the feed-forward width follows $\mathrm{MLP}=4d$. 
In parallel, we apply a mild scaling to the backbone by increasing both the number of blocks and its internal channel size ($bb\_dim$) to avoid an early bottleneck. 
Across all scales, tokenization (joint tokens $J$ plus point tokens) and training recipe are kept fixed, with RMSNorm and dropout ($p=0.1$) for stability.

\nosection{Scaling Grid}
Our study covers depths $D\in\{2,3,5,8,12,16\}$, heads $\in\{3,4,6,8,10,12\}$ (widths $d\in\{384,512,640,768\}$), and backbone blocks $\in\{2,3,4,5,6,8\}$ with $bb\_dim$ increasing from 32 to 192. 
For each configuration, we also log parameter counts and FLOPs to relate model size to performance to plot \figref{fig:size_vs_mpjpe}.

\begin{figure*}[t]
    \centering
    \rule{\linewidth}{0.8pt}
    \vspace{-2.5mm}
    
    \includegraphics[width=\linewidth]{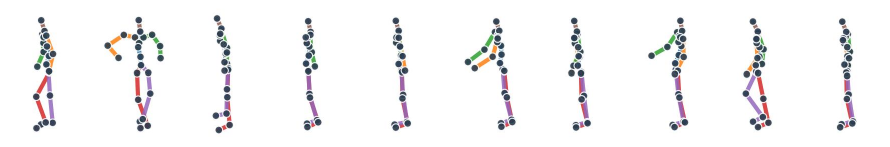}
    \vspace{-3mm}
    
    \includegraphics[width=\linewidth]{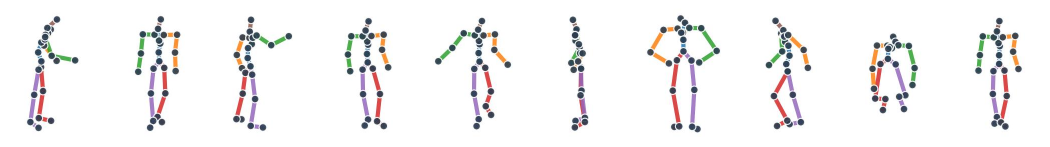}
    
    \vspace{-1mm}
    \rule{\linewidth}{0.8pt}
    \vspace{-1.5mm}
    
    \includegraphics[width=\linewidth]{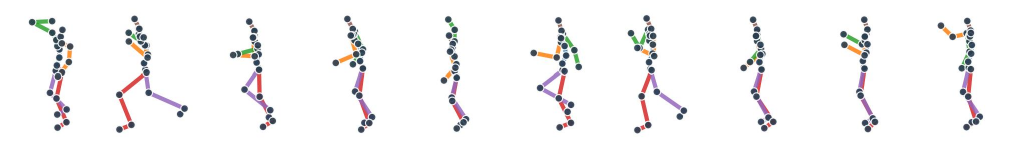}
    \vspace{-3mm}
    
    \includegraphics[width=\linewidth]{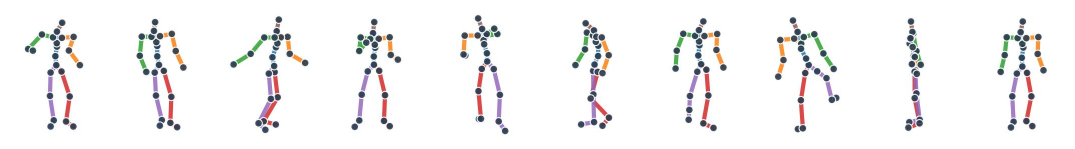}
    \vspace{-3mm}
    
    \vspace{-2.5mm}
    \rule{\linewidth}{0.8pt}
    \vspace{-5mm}
    
    \caption{Pose visualizations (randomly selected). From top to bottom: 
    (1) mmBody poses, 
    (2) poses in AMASS but uncovered by mmBody ($k = 10$), 
    (3) MMFi poses, and 
    (4) poses in AMASS but uncovered by MMFi ($k = 12$).}
    \vspace{-2mm}
    \label{fig:pose_visualizations}
\end{figure*}

\section{NRI Algorithm}\label{app:nri}

\nosection{Motivation}
The raw $k$-NN radius $r_{\phi}^{(k)}$ is a convenient, non-parametric proxy for local density, but it is not directly comparable across datasets because its magnitude depends on both the dataset size $n$ and the neighborhood size $k$. We therefore employ a \emph{scale-free} redundancy measure that preserves the geometric intuition of $r_{\phi}^{(k)}$ while removing these extrinsic dependencies.

\nosection{Definition}
For a dataset $\Phi=\{\phi_i\}_{i=1}^n$ embedded in the shared latent space (same encoder and metric as in the main text), the \emph{Normalized Redundancy Index} (NRI) of sample $\phi_i$ at scale $k$ is
\[
\mathrm{NRI}_i(k)=\frac{(n-1)\,[r_{i}^{(k)}]^{\,d_{\mathrm{eff}}}}{k},
\]
where $r_{i}^{(k)}$ is the distance from $\phi_i$ to its $k$-th nearest neighbor in $\Phi\setminus\{\phi_i\}$, and $d_{\mathrm{eff}}$ is the effective intrinsic dimension of the latent space. Operationally, $d_{\mathrm{eff}}$ is estimated from the slope of a robust regression between $\log r^{(k)}$ and $\log k$ across multiple $k$ values (median-based trimming), following the small-ball scaling $r^{(k)}\propto (k/n)^{1/d}$ \cite{levina2004maximum,facco2017estimating}.

\nosection{Why NRI is Scale Free}
Classical $k$-NN density arguments state that, in a small neighborhood around $\phi$,
\[
\mathbb{E}[\#\text{NN within }r]\approx (n-1)\,f(\phi)\,C_{d}\,r^{d},
\]
where $f(\phi)$ is the local density and $C_d$ is the unit-ball constant under the chosen metric. Solving for the radius at a fixed neighbor count $k$ gives
$
r_{\phi}^{(k)}\approx \big(k/[(n-1)\,f(\phi)\,C_d]\big)^{1/d}.
$
Substituting this into the definition above yields
\[
\mathrm{NRI}_\phi(k)\ \approx\ \frac{1}{f(\phi)\,C_d},
\]
which is \textit{independent of $n$ and (to first order) $k$}. In other words, NRI is a monotone transform of the standard $k$-NN density estimator \cite{loftsgaarden1965nonparametric,biau2015lectures}: smaller NRI indicates higher density and, consequently, higher redundancy.

\nosection{Implementation Details}
We compute $r_i^{(k)}$ for a set of scales $k$ (or, equivalently, for fixed coverage rates $\alpha=k/(n-1)$) in the same latent space~\cite{levina2004maximum,facco2017estimating} and metric as the main text. To avoid numerical issues under duplicates, we floor radii by a small $\varepsilon$ before taking logarithms. We estimate $d_{\mathrm{eff}}$ via a robust linear fit of $\log r^{(k)}$ versus $\log k$ (with mild trimming of extreme $k$ if needed), and then evaluate $\mathrm{NRI}_i(k)$ per sample.

\section{Uncovered Cases of mmWave HPE Datasets}\label{app:poses}
To qualitatively illustrate the coverage gaps identified by our cross-dataset analysis, we visualize several poses from the \textit{gap set} ($\Phi_{G}$). These examples, shown in \figref{fig:pose_visualizations}, represent poses that are present in the comprehensive AMASS dataset but are underrepresented in existing mmWave datasets. This visualization provides a concrete illustration of the abstract concept of coverage gaps and intuitively demonstrates the kind of informative data our guided collection strategy aims to acquire.

\newpage

\end{document}